\documentclass[11pt]{article}

\usepackage[final]{acl}

\usepackage{times}
\usepackage{latexsym}
\usepackage[T1]{fontenc}
\usepackage[utf8]{inputenc}
\usepackage{microtype}
\usepackage{inconsolata}
\usepackage{graphicx}

\usepackage{amsmath}
\usepackage{amssymb}
\usepackage{booktabs}
\usepackage{array}
\usepackage{tabularx}
\usepackage{multirow}
\usepackage{xcolor}
\usepackage{url}
\usepackage{algorithm}
\usepackage{algpseudocode}
\usepackage{enumitem}
\usepackage[most]{tcolorbox}
\tcbuselibrary{breakable,skins}

\newtcolorbox{promptbox}[1][]{
  enhanced,
  colback=gray!6,
  colframe=black!55,
  fonttitle=\bfseries\small,
  title={#1},
  arc=3pt,
  boxrule=0.5pt,
  left=6pt, right=6pt, top=4pt, bottom=6pt,
  parbox=false,
  attach boxed title to top left={yshift=-2mm, xshift=4mm},
  boxed title style={colback=black!55, colframe=black!55, arc=2pt,
                     boxrule=0pt, fontupper=\color{white}\bfseries\small},
}

\newcolumntype{Y}{>{\raggedright\arraybackslash}X}

\title{Do VLMs Align Better with Humans than LLMs Do during Natural Reading?}

\author{
  \textbf{Jinzhou Wu}\textsuperscript{1,2,5} \quad
  \textbf{Zhengwu Ma}\textsuperscript{3} \quad
  \textbf{Jixing Li}\textsuperscript{3} \quad
  \textbf{Baoping Tang}\textsuperscript{1} \quad
  \textbf{Zitong Lu}\textsuperscript{4}$^{*}$
\\
\\
  \textsuperscript{1}Department of Mechanical and Vehicle Engineering, Chongqing University, China \\
  \textsuperscript{2}College of Computing and Data Science, Nanyang Technological University, Singapore \\
  \textsuperscript{3}Department of Linguistics and Translation, City University of Hong Kong, China \\
  \textsuperscript{4}McGovern Institute for Brain Research, Massachusetts Institute of Technology, USA \\
  \textsuperscript{5}Sovi.ai, Edgewise Pte Ltd, Singapore
\\
  \small{
    $^{*}$\textbf{Correspondence:} \href{mailto:zitonglu@mit.edu}{zitonglu@mit.edu}
  }
}

\begin{document}
\maketitle

\begin{abstract}
Large language models (LLMs) have become increasingly useful computational models of human language processing, yet it remains open whether vision--language learning is associated with more human-like text representations during natural reading. 
We address this question by comparing matched LLM and vision--language model (VLM) pairs under strictly text-only input and evaluating alignment with human brain activity (whole-cortex fMRI) and reading behavior (synchronized regressive saccades). Across evaluated pairs, we identify a selective, rather than global, pattern in human--model alignment. In first-party continued-pretraining pairs (Qwen2 and DeepSeek), VLMs more accurately predicted human regressive saccades while exhibiting practical equivalence with matched LLMs in whole-cortex fMRI alignment, whereas a LoRA-adapted pair (Mistral/Idefics2) showed no divergence. However, sentence-level fMRI analyses revealed a growing VLM advantage for more visually evocative sentences. Together, these findings provide a controlled comparison of matched LLMs and VLMs, showing selective human-alignment advantages in reading behavior and visually grounded content.\footnote{Code is available at \url{https://github.com/wjz114/VLM-LLM-Brain}.}
\end{abstract}

\section{Introduction}

Large language models (LLMs) are increasingly used as computational models of human language processing, demonstrating a remarkable capacity for predicting neural activity and behavioral signals during natural reading, story listening, and language comprehension \citep{schrimpfNeuralArchitectureLanguage2021, goldsteinSharedComputationalPrinciples2022, alkhamissiLanguageCognitionHow2025}. Consequently, model alignment with human neural and behavioral responses has emerged as an increasingly important benchmark for evaluating whether language models capture human-like language processing \citep{caucheteuxBrainsAlgorithmsPartially2022}. A truly human-like natural language processing (NLP) model should not only perform well on downstream benchmarks but also mirror the mechanistic ways in which humans process inputs \citep{cichyDeepNeuralNetworks2019,lu2025teaching,yinImproveLanguageModel2025,lopez-cardonaBrainLanguageModel2026}.

However, existing human alignment research has largely focused on pure text representations \citep{merlinLanguageModelsBrains2024}, scaling laws \citep{gaoIncreasingAlignmentLarge2025}, next-word prediction objectives \citep{yuPredictingNextSentence2024a}, or instruction tuning \citep{ootaInstructionTunedVideoAudioModels2025}. From an AI perspective, the rapid development of vision-language models (VLMs) raises a natural question for model-human alignment: does multimodal pretraining make language representations more human-like, even when models process text alone? Unlike LLMs trained primarily on text, VLMs are exposed to paired visual and linguistic data, which may encourage representations that capture visually grounded aspects of meaning \citep{jonesMultimodalLargeLanguage2024a}. From a cognitive perspective, human language comprehension may also be shaped by multimodal sensorimotor experience, and many concepts are partly grounded in perceptual and action systems \citep{barsalouGroundedCognitionPresent2010,binderNeurobiologySemanticMemory2011, lynottLancasterSensorimotorNorms2020}. If visual experience contributes to human language processing during reading, VLMs may align more closely than matched LLMs with human neural and behavioral responses \citep{bavarescoModellingMultimodalIntegration2025, xuLargeLanguageModels2025}. 
This raises the central question of our study: whether and where vision-language learning influences model-human alignment during natural reading.
While NLP models cannot be viewed as literal replacements for human subjects \citep{cichyDeepNeuralNetworks2019}, matched LLM/VLM pairs offer a controlled \emph{in silico} system for addressing this question.

Empirical evidence on the comparative brain alignment of multimodal models remains mixed.
Concept-word brain alignment can favor VLMs \citep{bavarescoModellingMultimodalIntegration2025}, whereas comparisons focused on experiential semantic information can favor language-only models \citep{bavarescoExperientialSemanticInformation2025}. Also, natural-reading work has established neural and eye-movement alignment for language models \citep{gaoIncreasingAlignmentLarge2025,kuribayashiPsychometricPredictivePower2024}, and other studies examine online audiovisual movie viewing \citep{subramaniamRevealingVisionLanguageIntegration2024a,ootaInstructionTunedVideoAudioModels2025}. What remains unclear is not whether models can align with natural reading at all, but whether multimodal training history improves that alignment under text-only inputs, and whether any such effect is whole-cortex, process-level, or content-conditioned. Model-level factors such as architecture, scale, training corpus, and post-training further complicate attribution. A more direct test, therefore, requires matched LLM/VLM pairs with identical text inputs, with these estimands explicitly stated and tested.

In this work, we compare matched generative VLMs and their base LLMs when processing identical text-only input during natural reading, isolating online visual input at inference time. We first quantify internal representational differences within matched LLM/VLM pairs, then evaluate their alignment with human regressive saccades and whole-cortex fMRI across three estimands: behavioral regressive-saccade alignment, whole-cortex fMRI alignment, and sentence-level visual-strength-conditioned fMRI alignment. This design turns the broad VLM-versus-LLM question into a controlled, estimand-resolved comparison rather than a single aggregate brain-alignment score.

Our contributions are as follows:
\begin{itemize}

    \item We develop an estimand-aware matched-pair framework for comparing LLMs and VLMs during natural reading. By contrasting text-only language processing in matched model pairs, this framework separates multimodal training history from the effects of online visual input.
    
    \item We show that in within-lineage continued-pretraining pairs, VLMs diverge in internal language representations and better predict human regressive saccades, yet remain practically equivalent to matched LLMs in whole-cortex fMRI alignment during natural reading.
    
    \item We introduce a continuous visual-strength moderation analysis of fMRI alignment and show that the VLM--LLM alignment difference increases with sentence visual strength. These findings indicate that VLM alignment advantages during natural reading emerge selectively in rereading behavior and visually grounded content rather than globally across the cortex.

\end{itemize}

\begin{figure*}[!t]
    \centering
    \includegraphics[width=\textwidth]{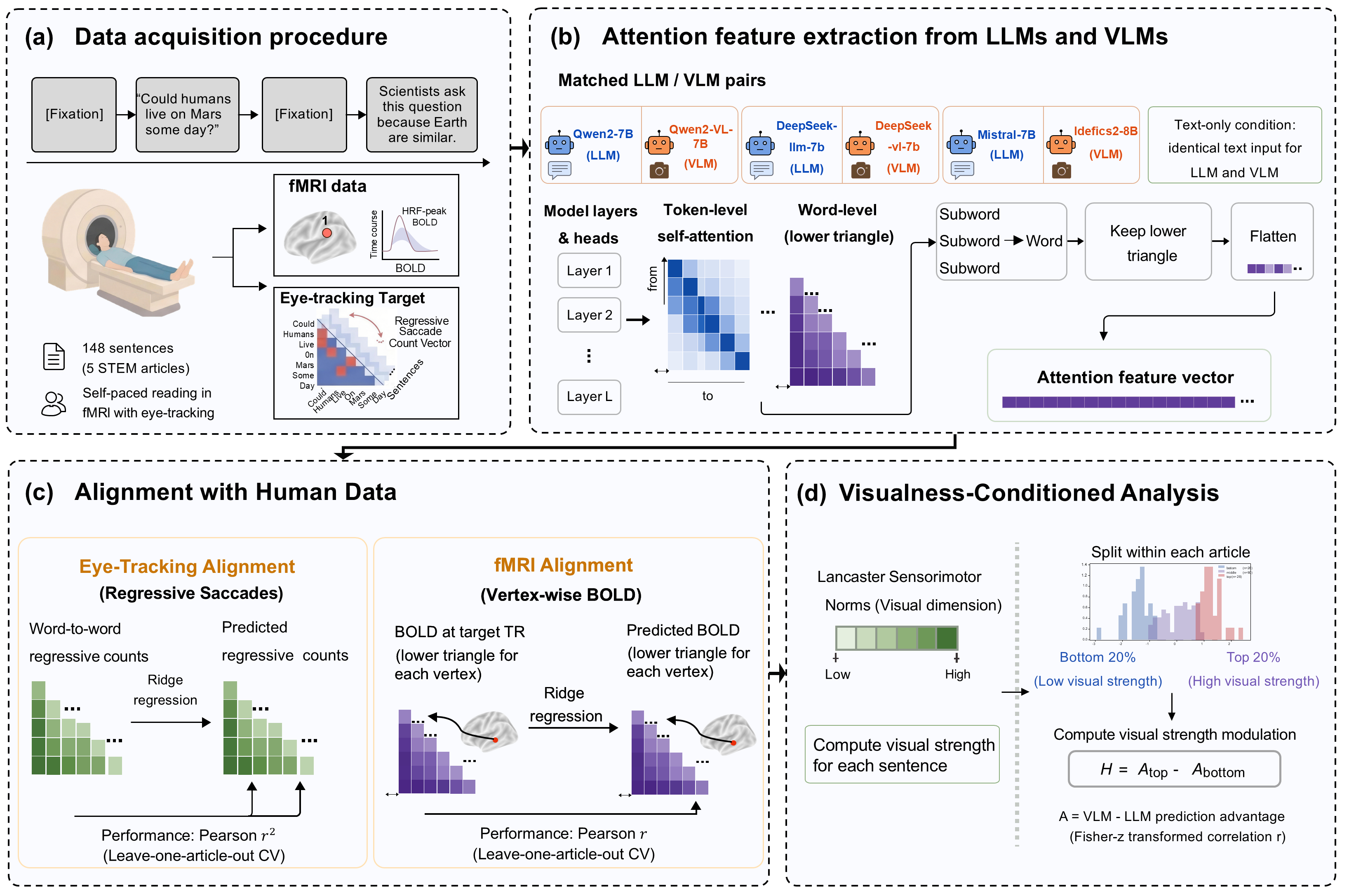}
    \caption{Overview of the matched LLM/VLM alignment pipeline. (a)~Data acquisition. Participants read sentences from five articles, one at a time, while concurrent eye-tracking and whole-cortex fMRI are recorded. (b)~Attention feature extraction. Layer-wise self-attention matrices are mapped from the token level to the word level, and the lower triangle is retained as the feature representation. (c)~Model-human alignment. Lower-triangular attention features are used as predictors in ridge regression models targeting vertex-wise fMRI responses and human regressive saccades during natural reading. (d)~A continuous visual-strength-moderation analysis regresses the sentence-level VLM--LLM fMRI-alignment difference on visual strength.}
    \label{fig:pipeline}
\end{figure*}

\section{Related Work}

Recent studies have increasingly used NLP models as computational probes of human language processing by comparing their internal representations with neural and behavioral responses \citep{schrimpfNeuralArchitectureLanguage2021,goldsteinSharedComputationalPrinciples2022,caucheteuxBrainsAlgorithmsPartially2022}. Transformer-based LLMs can predict fMRI, MEG, ECoG, and eye-movement signals during naturalistic language comprehension, suggesting shared computational structure between language models and the human language system \citep{tuckuteDrivingSuppressingHuman2024,antonelloScalingLawsLanguage2023,kuribayashiPsychometricPredictivePower2024}. Subsequent studies have examined which model properties drive this alignment, including model scale, predictive objectives, associative memory, instruction tuning, and intermediate-layer representations \citep{awInstructiontuningAlignsLLMs2024,gaoIncreasingAlignmentLarge2025}. These findings establish model--human alignment as a useful framework for evaluating whether NLP models capture human-like language processing. 

A related line of work asks whether multimodal experience yields more human-aligned language representations, with mixed conclusions across model architectures and experimental settings. Studying isolated concept words in picture or sentence contexts, \citet{bavarescoModellingMultimodalIntegration2025} report stronger VLM brain alignment overall, especially when visual context is present, whereas \citet{bavarescoExperientialSemanticInformation2025} find language-only models competitive with or superior to contrastive multimodal models on experiential semantic norms and concept-level fMRI. Other studies examine online audiovisual movie viewing \citep{subramaniamRevealingVisionLanguageIntegration2024a,ootaInstructionTunedVideoAudioModels2025} or sensorimotor and cross-modal concept representations \citep{xuLargeLanguageModels2025,ryskinaLanguageModelsAlign2025}. Our study complements these concept- and audiovisual-level analyses by examining continuous natural sentence reading under strictly text-only inputs in matched generative LLM/VLM pairs, and by jointly modeling concurrent regressive saccades and whole-cortex fMRI to separate global neural equivalence, rereading behavior, and sentence-level visual-strength moderation (Table~\ref{tab:estimand-matrix}, Appendix~\ref{app:related-work-matrix}).

Classical embedding-alignment paradigms map sequential hidden states to neural responses \citep{tonevaInterpretingImprovingNaturallanguage2019}. Our fixation-linked blood-oxygen-level-dependent (BOLD) and regressive-saccade targets instead motivate directed word-pair predictors. We adapt this natural-reading encoding setup \citep{gaoIncreasingAlignmentLarge2025} to matched LLM/VLM pairs under identical text-only inputs, jointly evaluating behavioral alignment, whole-cortex neural alignment, and content-conditioned neural differences.

\section{Methods}
We compared matched LLM/VLM pairs under text-only inference and evaluated their alignment with fMRI responses and regressive eye movements during natural reading, illustrated in Figure~\ref{fig:pipeline}. 

\subsection{Natural Reading Dataset}

We used the openly available Reading Brain dataset on OpenNeuro, which contains eye-tracking and fMRI recordings from 52 adult native English speakers reading naturalistic English texts \citep{liReadingComprehensionL12019}. Participants read 148 sentences from five English STEM articles, one sentence at a time, inside the scanner, while fMRI and eye movements were recorded concurrently. The data preprocessing details are in Appendix~\ref{app:model-details}.

\subsection{Matched LLM/VLM Pairs}

We evaluated three generative LLM/VLM pairs with matched language backbones at a comparable scale: (1) Qwen2-7B vs.\ Qwen2-VL-7B; (2) DeepSeek-LLM-7B vs.\ DeepSeek-VL-7B; and (3) Mistral-7B-v0.1 vs.\ Idefics2-8B-base. In the two first-party continued-pretraining pairs (Qwen2 and DeepSeek), the VLM is released by the same developer via full-backbone multimodal continued pretraining from the matched LLM lineage \citep{yangQwen2TechnicalReport2024,wangQwen2VLEnhancingVisionLanguage2024,deepseek-aiDeepSeekLLMScaling2024,luDeepSeekVLRealWorldVisionLanguage2024}. Idefics2-8B-base instead initializes from Mistral-7B-v0.1 in a third-party pipeline with a SigLIP vision encoder and Perceiver resampler, updating the pretrained backbones via low-rank adaptation (LoRA) while fully training newly initialized connector parameters \citep{jiangMistral7B2023,laurenconWhatMattersWhen2024}. We also evaluated a smaller within-family pair (Qwen2-1.5B vs.\ Qwen2-VL-2B) for scale sensitivity (Appendix~\ref{app:scale-sensitivity}).

All VLMs were run in text-only mode with no image or visual placeholder supplied, so that both models in each pair processed identical textual stimuli. None of these models undergoes instruction tuning. Detailed model configurations are listed in Appendix~\ref{app:model-details}.

\subsection{Attention Feature Extraction}

We follow a fixation-linked natural-reading encoding setup by  \citet{gaoIncreasingAlignmentLarge2025}. Each sentence was passed to each model independently, matching the experimental design in which participants saw one sentence at a time. For every sentence, model layer, and attention head, we extracted token-level self-attention matrices from the language model component. Because the models use subword tokenization, token-level attention was converted to word-level attention using the tokenizer's word IDs. Attention assigned to subword tokens belonging to the same source word was summed across columns, and attention from subword tokens belonging to the same target word was averaged across rows, yielding one word-by-word attention matrix per sentence.

Regressive saccades and fixation-linked BOLD responses are defined on directed word pairs. We therefore retain the strict lower triangle of each word-level matrix, yielding 7{,}421 shared word-pair entries across sentences. Attention supplies predictors in this space directly, and the contextual-hidden-state control is transformed into the same space. Feature construction details are given in Appendix~\ref{app:encoding-details}.

\subsection{Within-Pair Representation Comparisons}

We compared word-level hidden-state geometry using representational similarity analysis (RSA) and attention distributions using Jensen--Shannon divergence (JSD) on the 148 stimulus sentences. RSA constructs sentence-level representational dissimilarity matrices (RDMs) using Pearson correlation distance ($1 - r_{\mathrm{Pearson}}$) between mean-pooled word hidden states and compares paired models via Spearman correlation across upper-triangular RDM entries, summarized by the median across layers. For attention, we report no-sink JSD, which removes sentence-initial attention mass that can dominate raw divergence without reflecting word-to-word patterns. These analyses quantify descriptive within-pair representational differences under identical text-only inputs. Full procedures and attention-sink analyses are provided in Appendix~\ref{app:repchange}.

\subsection{Eye-Movement Alignment}

We tested whether model attention predicted human regressive saccade counts during sentence reading, which index rereading and integration of earlier context \citep{cliftonEyeMovementsReading2007}.

For each matched model pair, the VLM and LLM were compared using a two-sided paired $t$-test across participants on one participant-level nested-CV alignment score per model. For each outcome, the three prespecified model-pair comparisons were adjusted using the Benjamini--Hochberg procedure. All eye-movement and fMRI encoding analyses used five-fold leave-one-article-out (LOAO) outer cross-validation. Preprocessing, ridge hyperparameter selection, and layer selection were confined to the training portion of each outer fold and the held-out article was used only for final scoring. Encoding details are provided in Appendix~\ref{app:encoding-details}.

\subsection{fMRI Alignment}

For fMRI alignment, we used the same lower-triangular attention predictors to predict BOLD responses associated with regressive fixation transitions. For each transition, we sampled the BOLD response at the estimated hemodynamic peak, 5 s after fixation onset, and constructed a vertex-wise response vector matched to the same 7,421 word-pair entries. Ridge models were evaluated using the same LOAO encoding procedure, with Pearson's $r$ between predicted and observed BOLD responses as the accuracy metric. Single-model significance was assessed by permutation testing. Matched LLM/VLM pairs were compared using participant-level paired tests on these nested-CV whole-cortex scores. Multiple-comparison correction was applied across the three prespecified model-pair contrasts, and cluster-based permutation testing on the cortical surface \citep{marisNonparametricStatisticalTesting2007}. A contextual-hidden-state control using the same targets, CV, and layer-selection policy is reported in Appendix~\ref{app:hidden-control}.

For whole-cortex VLM--LLM contrasts, we distinguish practical equivalence from a nonsignificant result using two one-sided tests (TOST) against a working $\pm0.01$ Fisher-$z$ region, alongside paired Bayes factors ($\mathrm{BF}_{01}$). 

\subsection{Visual-Strength-Conditioned fMRI Analysis}

If visual experience selectively shapes the processing of visually grounded language, the VLM--LLM alignment difference may covary with how visually evocative a sentence is \citep{xuLargeLanguageModels2025}. We therefore tested whether VLMs show a greater fMRI-alignment advantage for more visually evocative sentences.

We defined sentence-level visual strength from human-rated norms as the mean Vision rating of matched content words in the Lancaster Sensorimotor Norms \citep{lynottLancasterSensorimotorNorms2020}, retaining 143/148 sentences with at least 75\% content-word coverage and at least three matched content words (a supplementary coverage-recovery check on low-coverage sentences is reported in Appendix~\ref{app:llm-fallback}).
Visual strength scores were standardized across these eligible sentences. For each participant $i$, cortical unit $u$, and sentence $s$, we computed the
out-of-fold VLM--LLM alignment difference in Fisher-$z$ space,
\begin{equation}
D_{ius} = z(r^{\mathrm{VLM}}_{ius}) -
          z(r^{\mathrm{LLM}}_{ius}).
\end{equation}
We then tested whether this difference increased with visual strength by regressing $D_{ius}$ on sentence visual strength:
\begin{equation}
\label{eq:moderation}
D_{ius} =
\alpha_{iu} + \beta_{iu} V_s +
\gamma_{iu} L_s + \varepsilon_{ius},
\end{equation}
where $V_s$ is standardized visual strength, $L_s$ is effective directed word-pair count (controlling for sample-size variation across sentences), $\beta_{iu}$ is the visual-strength moderation slope, $\alpha_{iu}$ is an intercept, $\gamma_{iu}$ captures the length effect, and $\varepsilon_{ius}$ is the residual. Parameters were fit by weighted least squares minimizing $\sum_{s} w_s (D_{ius} - \alpha_{iu} - \beta_{iu} V_s - \gamma_{iu} L_s)^2$, with sample weight $w_s$ set to sentence word count. A positive $\beta_{iu}$ means that the VLM--LLM alignment difference increases with visual strength (specification checks and explained variance are in Appendix~\ref{app:visualness-continuous}).

To display the relationship at the extremes of the visual strength
distribution, we additionally summarized the pattern using grouped visual strength
sentences. Within each article, the top and bottom 20\% of eligible sentences by visual strength score were labeled high and low visual strength. For each cortical unit $u$ and group
$g\in\{\mathrm{high},\mathrm{low}\}$, we computed
\begin{equation}
A_{u,g} = z(r^{\mathrm{VLM}}_{u,g}) -
          z(r^{\mathrm{LLM}}_{u,g}),
\end{equation}
and summarized the extreme-group contrast as
\begin{equation}
H_u = A_{u,\mathrm{high}} - A_{u,\mathrm{low}}.
\end{equation}
Additional threshold sensitivity analyses are reported in
Appendix~\ref{app:visualness-details}.

\section{Results}

\subsection{Within-Pair Models Differ in Both Representation Channels}

\begin{figure}[t]
    \centering
    \includegraphics[width=\columnwidth]{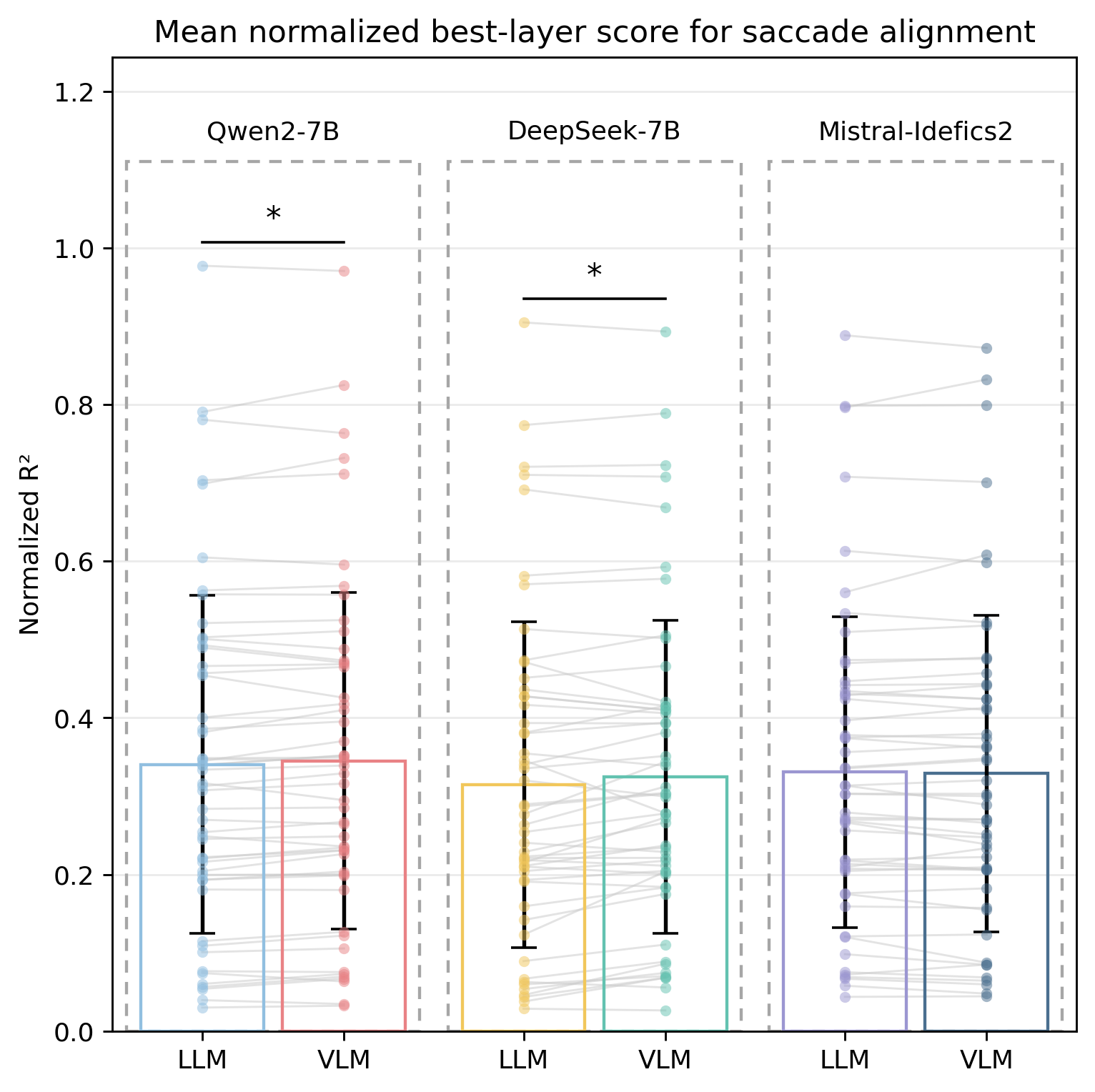}
    \caption{Eye-movement alignment for regressive saccades. Points show subject-level normalized out-of-sample $R^2$ values. Lines connect the matched LLM and VLM scores for each subject. Asterisks represent significant LLM--VLM differences within a pair on two-sided paired $t$-tests.}
    \label{fig:saccade}
\end{figure}

Before interpreting human-alignment contrasts, we quantified how the members of each pair differed under identical text-only inputs (Table~\ref{tab:representation-comparisons}). Under hidden-state RSA, all three matched pairs were substantially closer than the unrelated Qwen2/Mistral baseline ($.574$), yet diverged from identity ($\mathrm{RSA}<1.0$). Across both representation channels, Mistral/Idefics2 remained closest to its parent LLM, exhibiting the highest hidden-state RSA ($.952$) and the lowest no-sink attention JSD ($.0064$). These descriptive comparisons verify that the released LLM and VLM checkpoints possess related but measurably distinct internal representations when processing identical text-only input.

\begin{table}[t]
\centering
\small
\setlength{\tabcolsep}{2pt}
\begin{tabular}{@{}l c c@{}}
\toprule
\textbf{Pair} & \textbf{RSA} & \textbf{JSD (no-sink)} \\
\midrule
Qwen2 / Qwen2-VL & .897 & .0189 \\
DeepSeek / DeepSeek-VL & .810 & .0197 \\
Mistral / Idefics2 & .952 & .0064 \\
Qwen2 / Mistral reference & .574 & -- \\
\bottomrule
\end{tabular}
\caption{Descriptive within-pair representation comparisons.}
\label{tab:representation-comparisons}
\end{table}

\subsection{VLMs Better Predict Regressive Saccades in the In-Lineage Pairs}

\begin{figure*}[t]
    \centering
    \includegraphics[width=\linewidth]{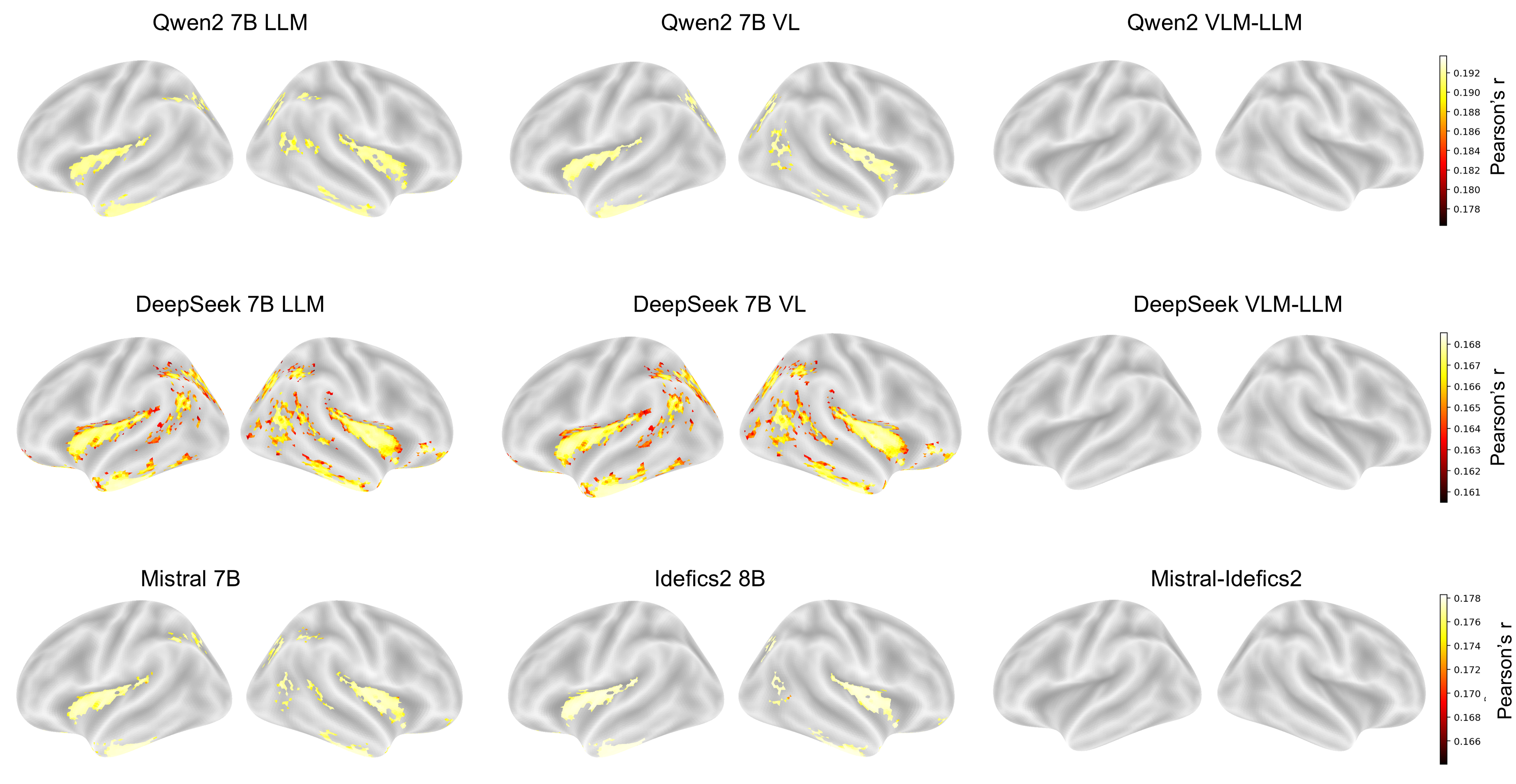}
    \caption{Global fMRI alignment of different LLM--VLM pairs. Significant brain clusters from the single-model-to-brain correlation coefficients and from the contrast of VLM to LLM. Colored regions denote vertex-wise mean effect sizes across subjects. The contrasts are small and do not support a broad whole-cortex fMRI advantage for VLMs during natural reading.}
    \label{fig:global-fmri}
\end{figure*}

We next tested whether VLM attention better predicted human regressive eye movements during natural reading. The eye-movement analysis revealed a modest, statistically reliable VLM advantage in both first-party continued-pretraining pairs, as shown in Figure~\ref{fig:saccade}.

Specifically, Qwen2-VL-7B outperformed Qwen2-7B in predicting regressive saccade counts under LOAO evaluation ($p{=}{.0072}$), and DeepSeek-VL-7B-base likewise outperformed DeepSeek-LLM-7B-base ($p{=}{.0020}$). By contrast, the Mistral/Idefics2 pair showed no statistically significant difference ($p{=}{.1711}$), mirroring its smaller within-pair representational divergence in Table~\ref{tab:representation-comparisons}.

\begin{figure*}[t]
    \centering
    \includegraphics[width=\linewidth]{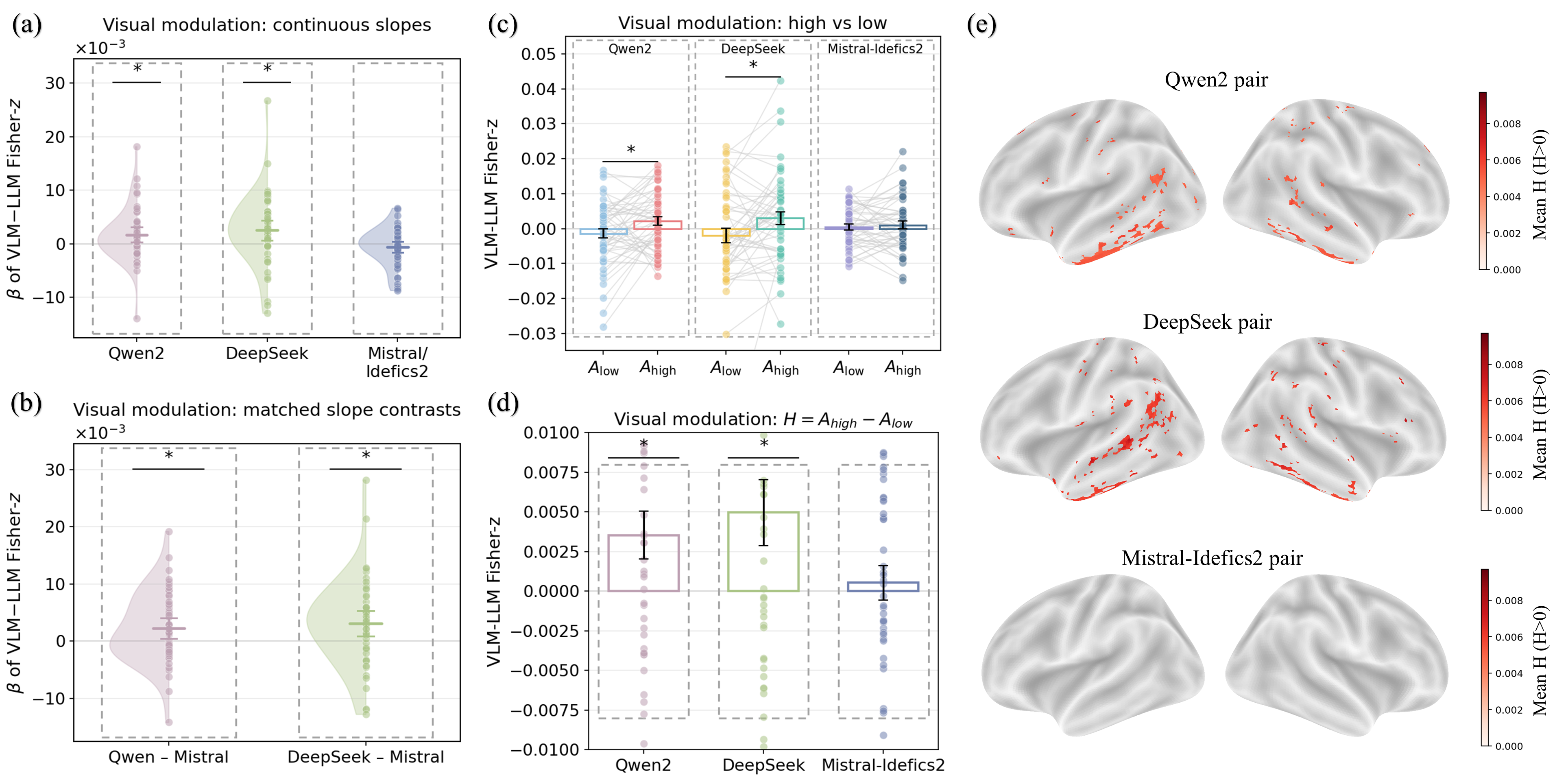}
    \caption{Visual strength modulation of VLM--LLM alignment across matched model pairs. (a)~Subject-level whole-cortex continuous moderation slopes $\beta$. Each dot is one subject and thick bars denote group means $\pm$ 95\% CI. Asterisks denote Holm-corrected one-sided sign-flip $p < .05$ across the three pairs. (b)~Within-subject slope contrasts against the cross-construction LoRA-adapted pair (Mistral/Idefics2), Holm-corrected across the two contrasts. (c)~Mean advantage in low- and high-visual-strength sentences. Bars denote group means $\pm$ SEM; dots denote individual subjects. (d)~Binary modulation index $H = A_{\mathrm{high}} - A_{\mathrm{low}}$ for each pair. Asterisks denote $p < .05$. (e)~Cortical localization of the high-vs-low visual strength contrast. Surface projections
show vertex-wise mean $H$ within clusters passing a
cluster-mass sign-flip permutation test, with cluster-forming $p = .01$,
cluster-level $\alpha = .05$ and 10{,}000 permutations.}
    \label{fig:visualness}
\end{figure*}

\subsection{Practical Equivalence in Whole-Cortex fMRI Alignment}

We then asked whether model attention predicted fMRI responses during text-only natural reading. Both LLMs and VLMs achieved reliable single-model fMRI alignment: surface-based significance maps revealed robust BOLD prediction across bilateral temporal--parietal cortex, encompassing canonical language and semantic networks (Figure~\ref{fig:global-fmri}).

Directly contrasting each VLM with its matched LLM, however, showed that whole-cortex VLM--LLM differences were practically equivalent within the working $\pm0.01$ Fisher-$z$ bound across all three pairs (Table~\ref{tab:fmri-equivalence}). Taken together with the eye-movement findings above, this establishes a clear estimand dissociation: the two full-backbone within-lineage pairs achieved higher VLM regressive-saccade alignment despite practically equivalent whole-cortex fMRI alignment. A contextual-hidden-state control independently reproduces both halves of this dissociation---whole-cortex fMRI equivalence across all three pairs and a selective VLM saccade advantage in the two first-party pairs (Appendix~\ref{app:hidden-control}). This pattern indicates selective, checkpoint-dependent alignment differences under text-only input rather than a uniform whole-brain VLM gain.

\begin{table}[t]
\centering
\small
\setlength{\tabcolsep}{2pt}
\begin{tabular}{@{}l c c c@{}}
\toprule
 \textbf{Pair} & $\Delta$ (95\% CI) & BF$_{01}$ & TOST \\
\midrule
 Qwen2 & $-$0.0006 $[{-}0.0033,0.0022]$ & 6.03 & Equiv. \\
 DeepSeek & $+$0.0005 $[{-}0.0021,0.0031]$ & 6.06 & Equiv. \\
 Mistral/Idefics2 & $+$0.0014 $[{-}0.0011,0.0038]$ & 3.60 & Equiv. \\
\bottomrule
\end{tabular}
\caption{Whole-cortex fMRI equivalence tests. Attention features were evaluated for all three pairs. $\Delta$ denotes the VLM--LLM difference in mean Fisher-$z$ alignment. Equiv.\ denotes equivalence under TOST with a working $\pm0.01$ Fisher-$z$ bound.}
\label{tab:fmri-equivalence}
\end{table}

\subsection{VLM--LLM fMRI Differences Scale with Visual Strength}
\label{sec:visualness-results}

A whole-cortex mean averages over all sentences and is insensitive to
content-dependent effects. We therefore estimated how the VLM--LLM
alignment difference scales with sentence-level visual strength.

Both first-party continued-pretraining pairs showed a positive
whole-cortex moderation slope that survived Holm correction, indicating
that their VLM--LLM fMRI difference grew with sentence visual strength
(Figure~\ref{fig:visualness}a;
Table~\ref{tab:visualness-continuous-appendix}). Visual strength accounted for an incremental $2.0\%$ (Qwen2) and $2.8\%$ (DeepSeek) of between-sentence alignment variance beyond article identity and sentence length, and slopes remained consistently positive under unweighted and article-fixed-effects specifications (Appendix~\ref{app:visualness-continuous}). By contrast, the
cross-construction Mistral/Idefics2 pair showed a numerically negative,
nonsignificant slope ($p{>}{.05}$), consistent with its null behavioral contrast. Participant-matched
slope contrasts confirmed that both in-lineage pairs had significantly larger
moderation slopes than Mistral/Idefics2 (Figure~\ref{fig:visualness}b).

The same three-way ordering is visible at the extremes of the visual-strength distribution (Figure~\ref{fig:visualness}c--d) and holds under a 25/25 split and exploratory coverage recovery
(Appendix~\ref{app:visualness-details}). Exploratory cortical
localization in Figure~\ref{fig:visualness}e places the positive moderation effect in bilateral lateral temporal cortex and the angular gyrus, regions associated with language and multimodal semantic integration \citep{ivanovaSemanticReasoningTakes2025, pophamVisualLinguisticSemantic2021}; given the modest effect magnitude, we treat this localization as descriptive. Finally, a within-family scale check on Qwen2-1.5B vs.\ Qwen2-VL-2B reproduced the VLM regressive-saccade advantage ($p{=}{.0207}$) and whole-cortex fMRI equivalence (TOST $p{<}{.001}$), whereas the visual-strength slope remained positive but nonsignificant ($p{=}{.266}$; Appendix~\ref{app:scale-sensitivity}).

\section{Discussion and Cognitive Implications}

Our representational analyses establish the model-side context for the human-alignment findings. Under identical text-only input, each matched LLM/VLM pair diverged measurably in hidden-state geometry and attention distributions, while varying across adaptation regimes. Notably, Idefics2 remained closely anchored to its Mistral-7B parent under both hidden-state RSA ($.952$) and no-sink attention JSD ($.0064$), and showed neither a regressive-saccade gain nor a positive visual-strength slope. This contrast highlights an adaptation-regime boundary condition: Idefics2 updates pretrained backbones via LoRA while training connector modules \citep{laurenconWhatMattersWhen2024}, preserving base-LLM text representations more strictly than the full-backbone continued pretraining in Qwen2-VL and DeepSeek-VL (consistent with \citealp{bavarescoModellingMultimodalIntegration2025}, who also found no significant Idefics2--Mistral advantage in sentence-only contexts). At the same time, because released VLMs combine multimodal supervision with additional text tokens and optimization updates, text-only inference isolates \emph{online visual input} rather than training-history confounds; the observed differences therefore reflect the composite VLM development history rather than visual exposure in isolation.

Across both first-party continued-pretraining pairs, VLMs better predicted regressive saccade counts---replicated with contextual hidden states (Appendix~\ref{app:hidden-control}). In reading research, regressive saccades index rereading, ambiguity resolution, and integration of earlier context with the current word \citep{cliftonEyeMovementsReading2007,raynerEyeMovementsAttention2009}. While our encoding model predicts directed word-pair return counts rather than labeling word-level semantic importance directly, this consistent VLM gain links multimodal adaptation to representations predictive of integrative rereading behavior.

Whole-cortex fMRI alignment provides the complementary global view. Across all evaluated pairs, VLMs and matched LLMs achieved practically equivalent whole-cortex BOLD prediction during text-only reading. Language-internal distributional, predictive, and compositional structure already accounts for a substantial share of whole-brain reading dynamics \citep{schrimpfNeuralArchitectureLanguage2021,goldsteinSharedComputationalPrinciples2022}, providing the common baseline on which content-conditioned differences emerge. Along the continuous Lancaster visual-strength dimension, VLM--LLM fMRI differences in the full-backbone pairs grew systematically---explaining an incremental $2.0\%$--$2.8\%$ of inter-sentence variance---with descriptive clusters in lateral temporal and inferior parietal regions implicated in multimodal conceptual integration \citep{ivanovaSemanticReasoningTakes2025,pophamVisualLinguisticSemantic2021}.

Beyond diagnostics, combining internal geometry (RSA and JSD), behavioral rereading alignment, and visual-strength slopes ($\beta$) offers two practical uses for model development: (1) \emph{model selection}, auditing checkpoints or adaptation regimes that enhance reading-process correspondence while preserving core language representations; and (2) \emph{objective alignment}, using rereading-predictive word-pair structure or content-conditioned neural correspondence as auxiliary regularizers or distillation targets in brain- and behavior-guided optimization \citep{shao2024probing,lu2025teaching,yinImproveLanguageModel2025,lu2026achieving}.

\section{Conclusion}

We compared three matched LLM/VLM pairs under strictly text-only inputs during natural reading across behavioral rereading alignment, whole-cortex fMRI alignment, and continuous visual-strength moderation. In both first-party continued-pretraining pairs (Qwen2 and DeepSeek), VLMs better predicted human regressive saccades and showed positive visual-strength moderation slopes, while remaining practically equivalent to matched LLMs in whole-cortex fMRI alignment; by contrast, LoRA-adapted Idefics2 remained close to Mistral-7B and showed no alignment divergence. Together, these findings show that estimand-aware matched-pair comparisons separate global neural equivalence from selective advantages in reading behavior and visually grounded language.

\section*{Limitations}

Our conclusions are bounded by several key limitations. First, the number of evaluated matched pairs remains small---three primary ${\sim}7$B--8B pairs and one smaller within-family Qwen2 pair---and is restricted to generative VLMs with identifiable autoregressive LLM parents. While the Qwen2-1.5B/2B evaluation shows that behavioral rereading gains and global fMRI equivalence are not unique to the 7B scale, the attenuation of the visual-strength moderation slope at 1.5B/2B underscores that content-dependent neural effects can vary across scales and families, limiting broad generalization without wider multi-family and multi-scale benchmarks.

Second, comparisons of publicly released LLM and VLM checkpoints inherently confound multimodal supervision with continued pretraining. Every VLM examined here was exposed to substantial additional text and multimodal tokens ($225$B--$1.4$T; Table~\ref{tab:model-configs}) alongside distinct data mixtures, connector modules, optimization schedules, and backbone update rules (full-parameter updates vs.\ LoRA). Strictly text-only inference eliminates online visual input at test time, but cannot causally disentangle visual grounding from continued text exposure or adaptation hyperparameters; doing so will require controlled from-scratch or matched-token continued-pretraining ablations.

Third, our strictly text-only reading paradigm does not capture real-time audiovisual or scene-grounded language comprehension, so global whole-cortex equivalence here does not imply that visual grounding is unimportant in multimodal environments. Finally, sentence-level visual strength derived from lexical Lancaster Vision norms serves only as an approximate bag-of-words proxy for compositional visual imagery and explains a modest fraction ($2.0\%$--$2.8\%$) of between-sentence variance. Future work should examine richer compositional imagery measures, broader eye-movement metrics beyond regressive saccade counts, and controlled training interventions.

\section*{Ethical Considerations}

This work aims to analyze the alignment between contemporary models and human language processing during natural reading. Such analyses may contribute to a better scientific understanding of both NLP systems and biological language processing, and may provide a biologically informed perspective on the interpretability and cognitive plausibility of current AI systems \citep{schrimpfNeuralArchitectureLanguage2021,gaoIncreasingAlignmentLarge2025}.

An openly available natural-reading dataset containing fMRI and eye-tracking recordings from adult participants \citep{liReadingComprehensionL12019} is used, and no new human-subject data were collected in this study. The original dataset was released for research use, and our analyses were conducted only on the de-identified public data. 

Our findings characterize aggregate model--human alignment in a specific natural-reading setting. By clarifying where current language and vision--language models do and do not align with human neural and behavioral responses, we hope this work contributes to the development of AI systems that are more interpretable, cognitively informed, and responsibly integrated into human-centered applications.

\section*{Acknowledgments}

This research was supported by the Innovation Research Group Project of National Natural Science Foundation of China under Grant T2421001 (to B.T.). At the same time, the authors gratefully acknowledge financial support from China Scholarship Council, China (to J.W.). We also thank the anonymous reviewers and the meta-reviewer for their constructive feedback.

\bibliography{custom}

@inproceedings{alkhamissiLanguageCognitionHow2025,
  title = {From {{Language}} to {{Cognition}}: {{How LLMs Outgrow}} the {{Human Language Network}}},
  shorttitle = {From {{Language}} to {{Cognition}}},
  booktitle = {Proceedings of the 2025 {{Conference}} on {{Empirical Methods}} in {{Natural Language Processing}}},
  author = {AlKhamissi, Badr and Tuckute, Greta and Tang, Yingtian and Binhuraib, Taha Osama A and Bosselut, Antoine and Schrimpf, Martin},
  editor = {Christodoulopoulos, Christos and Chakraborty, Tanmoy and Rose, Carolyn and Peng, Violet},
  year = {2025},
  month = nov,
  pages = {24321--24339},
  publisher = {Association for Computational Linguistics},
  address = {Suzhou, China},
  doi = {10.18653/v1/2025.emnlp-main.1237},
  urldate = {2026-05-24},
  isbn = {9798891763326},
  langid = {english}
}

@inproceedings{antonelloScalingLawsLanguage2023,
  title = {Scaling Laws for Language Encoding Models in {{fMRI}}},
  booktitle = {Thirty-Seventh {{Conference}} on {{Neural Information Processing Systems}}},
  author = {Antonello, Richard and Vaidya, Aditya and Huth, Alexander},
  year = {2023},
  month = nov,
  urldate = {2026-05-24},
  langid = {english}
}

@inproceedings{awInstructiontuningAlignsLLMs2024,
  title = {Instruction-Tuning {{Aligns LLMs}} to the {{Human Brain}}},
  booktitle = {First {{Conference}} on {{Language Modeling}}},
  author = {Aw, Khai Loong and Montariol, Syrielle and AlKhamissi, Badr and Schrimpf, Martin and Bosselut, Antoine},
  year = {2024},
  month = aug,
  urldate = {2026-05-24},
  langid = {english}
}

@article{barsalouGroundedCognitionPresent2010,
  title = {Grounded {{Cognition}}: {{Past}}, {{Present}}, and {{Future}}},
  shorttitle = {Grounded {{Cognition}}},
  author = {Barsalou, Lawrence W.},
  year = {2010},
  journal = {Topics in Cognitive Science},
  volume = {2},
  number = {4},
  pages = {716--724},
  issn = {1756-8765},
  doi = {10.1111/j.1756-8765.2010.01115.x},
  urldate = {2026-05-24},
  copyright = {Copyright {\copyright} 2010 Cognitive Science Society, Inc.},
  langid = {english}
}

@misc{bavarescoExperientialSemanticInformation2025,
  title = {Experiential {{Semantic Information}} and {{Brain Alignment}}: {{Are Multimodal Models Better}} than {{Language Models}}?},
  shorttitle = {Experiential {{Semantic Information}} and {{Brain Alignment}}},
  author = {Bavaresco, Anna and Fern{\'a}ndez, Raquel},
  year = {2025},
  month = jun,
  number = {arXiv:2504.00942},
  eprint = {2504.00942},
  primaryclass = {cs},
  publisher = {arXiv},
  doi = {10.48550/arXiv.2504.00942},
  urldate = {2025-12-22},
  archiveprefix = {arXiv},
  langid = {english}
}

@misc{bavarescoModellingMultimodalIntegration2025,
  title = {Modelling {{Multimodal Integration}} in {{Human Concept Processing}} with {{Vision-Language Models}}},
  author = {Bavaresco, Anna and Kloots, Marianne de Heer and Pezzelle, Sandro and Fern{\'a}ndez, Raquel},
  year = {2025},
  month = apr,
  number = {arXiv:2407.17914},
  eprint = {2407.17914},
  primaryclass = {cs},
  publisher = {arXiv},
  doi = {10.48550/arXiv.2407.17914},
  urldate = {2026-01-08},
  archiveprefix = {arXiv},
  langid = {english}
}

@article{binderNeurobiologySemanticMemory2011,
  title = {The {{Neurobiology}} of {{Semantic Memory}}},
  author = {Binder, Jeffrey R. and Desai, Rutvik H.},
  year = {2011},
  month = nov,
  journal = {Trends in Cognitive Sciences},
  volume = {15},
  number = {11},
  pages = {527--536},
  issn = {1364-6613},
  doi = {10.1016/j.tics.2011.10.001},
  urldate = {2026-05-24},
  langid = {english},
  lccn = {1},
  pmcid = {PMC3350748},
  pmid = {22001867}
}

@article{caucheteuxBrainsAlgorithmsPartially2022,
  title = {Brains and Algorithms Partially Converge in Natural Language Processing},
  author = {Caucheteux, Charlotte and King, Jean-R{\'e}mi},
  year = {2022},
  month = feb,
  journal = {Communications Biology},
  volume = {5},
  number = {1},
  pages = {134},
  issn = {2399-3642},
  doi = {10.1038/s42003-022-03036-1},
  urldate = {2026-05-24},
  langid = {english},
  lccn = {2}
}

@article{cichyDeepNeuralNetworks2019,
  title = {Deep {{Neural Networks}} as {{Scientific Models}}},
  author = {Cichy, Radoslaw M. and Kaiser, Daniel},
  year = {2019},
  month = apr,
  journal = {Trends in Cognitive Sciences},
  volume = {23},
  number = {4},
  pages = {305--317},
  issn = {13646613},
  doi = {10.1016/j.tics.2019.01.009},
  urldate = {2026-05-24},
  langid = {english},
  lccn = {1}
}

@incollection{cliftonEyeMovementsReading2007,
  title = {Eye Movements in Reading Words and Sentences},
  booktitle = {Eye {{Movements}}},
  author = {Clifton, Charles and Staub, Adrian and Rayner, Keith},
  year = {2007},
  month = dec,
  pages = {341--371},
  publisher = {Elsevier},
  isbn = {978-0-08-044980-7},
  langid = {english}
}

@article{gaoIncreasingAlignmentLarge2025,
  title = {Increasing Alignment of Large Language Models with Language Processing in the Human Brain},
  author = {Gao, Changjiang and Ma, Zhengwu and Chen, Jiajun and Li, Ping and Huang, Shujian and Li, Jixing},
  year = {2025},
  month = sep,
  journal = {Nature Computational Science},
  volume = {5},
  number = {11},
  pages = {1080--1090},
  issn = {2662-8457},
  doi = {10.1038/s43588-025-00863-0},
  urldate = {2025-12-02},
  langid = {english}
}

@article{goldsteinSharedComputationalPrinciples2022,
  title = {Shared Computational Principles for Language Processing in Humans and Deep Language Models},
  author = {Goldstein, Ariel and Zada, Zaid and Buchnik, Eliav and Schain, Mariano and Price, Amy and Aubrey, Bobbi and Nastase, Samuel A. and Feder, Amir and Emanuel, Dotan and Cohen, Alon and Jansen, Aren and Gazula, Harshvardhan and Choe, Gina and Rao, Aditi and Kim, Catherine and Casto, Colton and Fanda, Lora and Doyle, Werner and Friedman, Daniel and Dugan, Patricia and Melloni, Lucia and Reichart, Roi and Devore, Sasha and Flinker, Adeen and Hasenfratz, Liat and Levy, Omer and Hassidim, Avinatan and Brenner, Michael and Matias, Yossi and Norman, Kenneth A. and Devinsky, Orrin and Hasson, Uri},
  year = {2022},
  month = mar,
  journal = {Nature Neuroscience},
  volume = {25},
  number = {3},
  pages = {369--380},
  issn = {1097-6256, 1546-1726},
  doi = {10.1038/s41593-022-01026-4},
  urldate = {2026-05-24},
  langid = {english},
  lccn = {1}
}

@misc{ivanovaSemanticReasoningTakes2025,
  title = {Semantic Reasoning Takes Place Largely Outside the Language Network},
  author = {Ivanova, Anna A. and Kauf, Carina and Gao, Ruimin and She, Jingyuan Selena and Kean, Hope H. and Goldhaber, Tanya and {Nieto-Casta{\~n}{\'o}n}, Alfonso and Varley, Rosemary and Kanwisher, Nancy and Fedorenko, Evelina},
  year = {2025},
  month = dec,
  primaryclass = {New Results},
  pages = {2025.12.07.692873},
  publisher = {bioRxiv},
  issn = {2692-8205},
  doi = {10.64898/2025.12.07.692873},
  urldate = {2026-05-23},
  archiveprefix = {bioRxiv},
  chapter = {New Results},
  copyright = {{\copyright} 2025, Posted by openRxiv. This pre-print is available under a Creative Commons License (Attribution 4.0 International), CC BY 4.0, as described at http://creativecommons.org/licenses/by/4.0/},
  langid = {english}
}

@article{jonesMultimodalLargeLanguage2024a,
  title = {Do {{Multimodal Large Language Models}} and {{Humans Ground Language Similarly}}?},
  author = {Jones, Cameron R. and Bergen, Benjamin and Trott, Sean},
  year = {2024},
  month = dec,
  journal = {Computational Linguistics},
  volume = {50},
  number = {4},
  pages = {1415--1440},
  issn = {0891-2017, 1530-9312},
  doi = {10.1162/coli_a_00531},
  urldate = {2025-12-23},
  langid = {english},
  lccn = {3}
}

@inproceedings{kuribayashiPsychometricPredictivePower2024,
  title = {Psychometric {{Predictive Power}} of {{Large Language Models}}},
  booktitle = {Findings of the {{Association}} for {{Computational Linguistics}}: {{NAACL}} 2024},
  author = {Kuribayashi, Tatsuki and Oseki, Yohei and Baldwin, Timothy},
  editor = {Duh, Kevin and Gomez, Helena and Bethard, Steven},
  year = {2024},
  month = jun,
  pages = {1983--2005},
  publisher = {Association for Computational Linguistics},
  address = {Mexico City, Mexico},
  doi = {10.18653/v1/2024.findings-naacl.129},
  urldate = {2026-05-23},
  langid = {english}
}

@inproceedings{lopez-cardonaBrainLanguageModel2026,
  title = {Brain--{{Language Model Alignment}}: {{Insights}} into the {{Platonic Hypothesis}} and {{Intermediate-Layer Advantage}}},
  shorttitle = {Brain--{{Language Model Alignment}}},
  booktitle = {Proceedings of {{UniReps}}: The {{Third Edition}} of the {{Workshop}} on {{Unifying Representations}} in {{Neural Models}}},
  author = {{Lopez-Cardona}, Angela and Idesis, Sebastian and Bruns, Mireia Masias and Abadal, Sergi and Arapakis, Ioannis},
  year = {2026},
  month = feb,
  pages = {81--101},
  publisher = {PMLR},
  issn = {2640-3498},
  urldate = {2026-05-23},
  langid = {english}
}

@article{lynottLancasterSensorimotorNorms2020,
  title = {The {{Lancaster Sensorimotor Norms}}: Multidimensional Measures of Perceptual and Action Strength for 40,000 {{English}} Words},
  shorttitle = {The {{Lancaster Sensorimotor Norms}}},
  author = {Lynott, Dermot and Connell, Louise and Brysbaert, Marc and Brand, James and Carney, James},
  year = {2020},
  month = jun,
  journal = {Behavior Research Methods},
  volume = {52},
  number = {3},
  pages = {1271--1291},
  issn = {1554-3528},
  doi = {10.3758/s13428-019-01316-z},
  langid = {english},
  pmcid = {PMC7280349},
  pmid = {31832879}
}

@article{marisNonparametricStatisticalTesting2007,
  title = {Nonparametric Statistical Testing of {{EEG-}} and {{MEG-data}}},
  author = {Maris, Eric and Oostenveld, Robert},
  year = {2007},
  month = aug,
  journal = {Journal of Neuroscience Methods},
  volume = {164},
  number = {1},
  pages = {177--190},
  issn = {0165-0270},
  doi = {10.1016/j.jneumeth.2007.03.024},
  urldate = {2026-05-24},
  langid = {english},
  lccn = {4}
}

@inproceedings{merlinLanguageModelsBrains2024,
  title = {Language Models and Brains Align Due to More than Next-Word Prediction and Word-Level Information},
  booktitle = {Proceedings of the 2024 {{Conference}} on {{Empirical Methods}} in {{Natural Language Processing}}},
  author = {Merlin, Gabriele and Toneva, Mariya},
  year = {2024},
  pages = {18431--18454},
  publisher = {Association for Computational Linguistics},
  address = {Miami, Florida, USA},
  doi = {10.18653/v1/2024.emnlp-main.1024},
  urldate = {2026-05-13},
  langid = {english}
}

@misc{ootaInstructionTunedVideoAudioModels2025,
  title = {Instruction-{{Tuned Video-Audio Models Elucidate Functional Specialization}} in the {{Brain}}},
  author = {Oota, Subba Reddy and Pahwa, Khushbu and Jindal, Prachi and Namburi, Satya Sai Srinath and Singh, Maneesh and Chakraborty, Tanmoy and Raju, Bapi S. and Gupta, Manish},
  year = {2025},
  month = jun,
  number = {arXiv:2506.08277},
  eprint = {2506.08277},
  primaryclass = {q-bio},
  publisher = {arXiv},
  doi = {10.48550/arXiv.2506.08277},
  urldate = {2025-11-28},
  archiveprefix = {arXiv},
  langid = {english}
}

@article{pophamVisualLinguisticSemantic2021,
  title = {Visual and Linguistic Semantic Representations Are Aligned at the Border of Human Visual Cortex},
  author = {Popham, Sara F. and Huth, Alexander G. and Bilenko, Natalia Y. and Deniz, Fatma and Gao, James S. and {Nunez-Elizalde}, Anwar O. and Gallant, Jack L.},
  year = {2021},
  month = nov,
  journal = {Nature Neuroscience},
  volume = {24},
  number = {11},
  pages = {1628--1636},
  publisher = {Nature Publishing Group},
  issn = {1546-1726},
  doi = {10.1038/s41593-021-00921-6},
  urldate = {2026-05-23},
  copyright = {2021 The Author(s), under exclusive licence to Springer Nature America, Inc.},
  langid = {english},
  lccn = {1}
}

@article{raynerEyeMovementsAttention2009,
  title = {Eye Movements and Attention in Reading, Scene Perception, and Visual Search},
  author = {Rayner, Keith},
  year = {2009},
  month = aug,
  journal = {Quarterly Journal of Experimental Psychology},
  volume = {62},
  number = {8},
  pages = {1457--1506},
  address = {2006},
  issn = {1747-0226},
  doi = {10.1080/17470210902816461},
  langid = {english},
  lccn = {4},
  pmid = {19449261}
}

@misc{ryskinaLanguageModelsAlign2025,
  title = {Language Models Align with Brain Regions That Represent Concepts across Modalities},
  author = {Ryskina, Maria and Tuckute, Greta and Fung, Alexander and Malkin, Ashley and Fedorenko, Evelina},
  year = {2025},
  month = aug,
  number = {arXiv:2508.11536},
  eprint = {2508.11536},
  primaryclass = {cs},
  publisher = {arXiv},
  doi = {10.48550/arXiv.2508.11536},
  urldate = {2026-01-08},
  archiveprefix = {arXiv},
  langid = {english}
}

@article{schrimpfNeuralArchitectureLanguage2021,
  title = {The Neural Architecture of Language: {{Integrative}} Modeling Converges on Predictive Processing},
  shorttitle = {The Neural Architecture of Language},
  author = {Schrimpf, Martin and Blank, Idan Asher and Tuckute, Greta and Kauf, Carina and Hosseini, Eghbal A. and Kanwisher, Nancy and Tenenbaum, Joshua B. and Fedorenko, Evelina},
  year = {2021},
  month = nov,
  journal = {Proceedings of the National Academy of Sciences},
  volume = {118},
  number = {45},
  pages = {e2105646118},
  issn = {0027-8424, 1091-6490},
  doi = {10.1073/pnas.2105646118},
  urldate = {2026-05-24},
  langid = {english}
}

@misc{subramaniamRevealingVisionLanguageIntegration2024a,
  title = {Revealing {{Vision-Language Integration}} in the {{Brain}} with {{Multimodal Networks}}},
  author = {Subramaniam, Vighnesh and Conwell, Colin and Wang, Christopher and Kreiman, Gabriel and Katz, Boris and Cases, Ignacio and Barbu, Andrei},
  year = {2024},
  month = jun,
  number = {arXiv:2406.14481},
  eprint = {2406.14481},
  primaryclass = {cs},
  publisher = {arXiv},
  doi = {10.48550/arXiv.2406.14481},
  urldate = {2025-12-04},
  archiveprefix = {arXiv},
  langid = {english}
}

@incollection{tonevaInterpretingImprovingNaturallanguage2019,
  title = {Interpreting and Improving Natural-Language Processing (in Machines) with Natural Language-Processing (in the Brain)},
  booktitle = {Proceedings of the 33rd {{International Conference}} on {{Neural Information Processing Systems}}},
  author = {Toneva, Mariya and Wehbe, Leila},
  year = {2019},
  month = dec,
  number = {1339},
  pages = {14954--14964},
  publisher = {Curran Associates Inc.},
  address = {Red Hook, NY, USA},
  urldate = {2026-05-24},
  langid = {english}
}

@article{tuckuteDrivingSuppressingHuman2024,
  title = {Driving and Suppressing the Human Language Network Using Large Language Models},
  author = {Tuckute, Greta and Sathe, Aalok and Srikant, Shashank and Taliaferro, Maya and Wang, Mingye and Schrimpf, Martin and Kay, Kendrick and Fedorenko, Evelina},
  year = {2024},
  month = mar,
  journal = {Nature Human Behaviour},
  volume = {8},
  number = {3},
  pages = {544--561},
  publisher = {Nature Publishing Group},
  issn = {2397-3374},
  doi = {10.1038/s41562-023-01783-7},
  urldate = {2026-05-24},
  copyright = {2024 The Author(s), under exclusive licence to Springer Nature Limited},
  langid = {english},
  lccn = {1}
}

@article{xuLargeLanguageModels2025,
  title = {Large Language Models without Grounding Recover Non-Sensorimotor but Not Sensorimotor Features of Human Concepts},
  author = {Xu, Qihui and Peng, Yingying and Nastase, Samuel A. and Chodorow, Martin and Wu, Minghua and Li, Ping},
  year = {2025},
  month = jun,
  journal = {Nature Human Behaviour},
  volume = {9},
  number = {9},
  pages = {1871--1886},
  issn = {2397-3374},
  doi = {10.1038/s41562-025-02203-8},
  urldate = {2026-05-23},
  langid = {english},
  lccn = {1}
}

@inproceedings{yinImproveLanguageModel2025,
  title = {Improve {{Language Model}} and {{Brain Alignment}} via {{Associative Memory}}},
  booktitle = {Findings of the {{Association}} for {{Computational Linguistics}}: {{ACL}} 2025},
  author = {Yin, Congchi and Zhang, Yongpeng and Wen, Xuyun and Li, Piji},
  editor = {Che, Wanxiang and Nabende, Joyce and Shutova, Ekaterina and Pilehvar, Mohammad Taher},
  year = {2025},
  month = jul,
  pages = {986--999},
  publisher = {Association for Computational Linguistics},
  address = {Vienna, Austria},
  doi = {10.18653/v1/2025.findings-acl.55},
  urldate = {2026-05-24},
  isbn = {9798891762565},
  langid = {english}
}

@article{yuPredictingNextSentence2024a,
  title = {Predicting the next Sentence (Not Word) in Large Language Models: {{What}} Model-Brain Alignment Tells Us about Discourse Comprehension},
  shorttitle = {Predicting the next Sentence (Not Word) in Large Language Models},
  author = {Yu, Shaoyun and Gu, Chanyuan and Huang, Kexin and Li, Ping},
  year = {2024},
  month = may,
  journal = {Science Advances},
  volume = {10},
  number = {21},
  pages = {eadn7744},
  publisher = {American Association for the Advancement of Science},
  doi = {10.1126/sciadv.adn7744},
  urldate = {2026-05-24},
  langid = {english},
  lccn = {1}
}

@misc{deepseek-aiDeepSeekLLMScaling2024,
  title = {{{DeepSeek LLM}}: {{Scaling Open-Source Language Models}} with {{Longtermism}}},
  shorttitle = {{{DeepSeek LLM}}},
  author = { Bi, Xiao and Chen, Deli and Chen, Guanting and Chen, Shanhuang and Dai, Damai and Deng, Chengqi and Ding, Honghui and Dong, Kai and Du, Qiushi and Fu, Zhe and Gao, Huazuo and Gao, Kaige and Gao, Wenjun and Ge, Ruiqi and Guan, Kang and Guo, Daya and Guo, Jianzhong and Hao, Guangbo and Hao, Zhewen and He, Ying and Hu, Wenjie and Huang, Panpan and Li, Erhang and Li, Guowei and Li, Jiashi and Li, Yao and Li, Y. K. and Liang, Wenfeng and Lin, Fangyun and Liu, A. X. and Liu, Bo and Liu, Wen and Liu, Xiaodong and Liu, Xin and Liu, Yiyuan and Lu, Haoyu and Lu, Shanghao and Luo, Fuli and Ma, Shirong and Nie, Xiaotao and Pei, Tian and Piao, Yishi and Qiu, Junjie and Qu, Hui and Ren, Tongzheng and Ren, Zehui and Ruan, Chong and Sha, Zhangli and Shao, Zhihong and Song, Junxiao and Su, Xuecheng and Sun, Jingxiang and Sun, Yaofeng and Tang, Minghui and Wang, Bingxuan and Wang, Peiyi and Wang, Shiyu and Wang, Yaohui and Wang, Yongji and Wu, Tong and Wu, Y. and Xie, Xin and Xie, Zhenda and Xie, Ziwei and Xiong, Yiliang and Xu, Hanwei and Xu, R. X. and Xu, Yanhong and Yang, Dejian and You, Yuxiang and Yu, Shuiping and Yu, Xingkai and Zhang, B. and Zhang, Haowei and Zhang, Lecong and Zhang, Liyue and Zhang, Mingchuan and Zhang, Minghua and Zhang, Wentao and Zhang, Yichao and Zhao, Chenggang and Zhao, Yao and Zhou, Shangyan and Zhou, Shunfeng and Zhu, Qihao and Zou, Yuheng},
  year = {2024},
  month = jan,
  number = {arXiv:2401.02954},
  eprint = {2401.02954},
  primaryclass = {cs.CL},
  publisher = {arXiv},
  doi = {10.48550/arXiv.2401.02954},
  urldate = {2026-05-24},
  archiveprefix = {arXiv},
  langid = {english}
}

@misc{jiangMistral7B2023,
  title = {Mistral {{7B}}},
  author = {Jiang, Albert Q. and Sablayrolles, Alexandre and Mensch, Arthur and Bamford, Chris and Chaplot, Devendra Singh and de las Casas, Diego and Bressand, Florian and Lengyel, Gianna and Lample, Guillaume and Saulnier, Lucile and Lavaud, L{\'e}lio Renard and Lachaux, Marie-Anne and Stock, Pierre and Scao, Teven Le and Lavril, Thibaut and Wang, Thomas and Lacroix, Timoth{\'e}e and Sayed, William El},
  year = {2023},
  month = oct,
  number = {arXiv:2310.06825},
  eprint = {2310.06825},
  primaryclass = {cs.CL},
  publisher = {arXiv},
  doi = {10.48550/arXiv.2310.06825},
  urldate = {2026-05-24},
  archiveprefix = {arXiv},
  langid = {english}
}

@misc{laurenconWhatMattersWhen2024,
  title = {What Matters When Building Vision-Language Models?},
  author = {Lauren{\c c}on, Hugo and Tronchon, L{\'e}o and Cord, Matthieu and Sanh, Victor},
  year = {2024},
  month = may,
  number = {arXiv:2405.02246},
  eprint = {2405.02246},
  primaryclass = {cs.CV},
  publisher = {arXiv},
  doi = {10.48550/arXiv.2405.02246},
  urldate = {2026-05-24},
  archiveprefix = {arXiv},
  langid = {english}
}

@misc{luDeepSeekVLRealWorldVisionLanguage2024,
  title = {{{DeepSeek-VL}}: {{Towards Real-World Vision-Language Understanding}}},
  shorttitle = {{{DeepSeek-VL}}},
  author = {Lu, Haoyu and Liu, Wen and Zhang, Bo and Wang, Bingxuan and Dong, Kai and Liu, Bo and Sun, Jingxiang and Ren, Tongzheng and Li, Zhuoshu and Yang, Hao and Sun, Yaofeng and Deng, Chengqi and Xu, Hanwei and Xie, Zhenda and Ruan, Chong},
  year = {2024},
  month = mar,
  number = {arXiv:2403.05525},
  eprint = {2403.05525},
  primaryclass = {cs.AI},
  publisher = {arXiv},
  doi = {10.48550/arXiv.2403.05525},
  urldate = {2026-05-24},
  archiveprefix = {arXiv},
  langid = {english}
}

@misc{wangQwen2VLEnhancingVisionLanguage2024,
  title = {Qwen2-{{VL}}: {{Enhancing Vision-Language Model}}'s {{Perception}} of the {{World}} at {{Any Resolution}}},
  shorttitle = {Qwen2-{{VL}}},
  author = {Wang, Peng and Bai, Shuai and Tan, Sinan and Wang, Shijie and Fan, Zhihao and Bai, Jinze and Chen, Keqin and Liu, Xuejing and Wang, Jialin and Ge, Wenbin and Fan, Yang and Dang, Kai and Du, Mengfei and Ren, Xuancheng and Men, Rui and Liu, Dayiheng and Zhou, Chang and Zhou, Jingren and Lin, Junyang},
  year = {2024},
  month = oct,
  number = {arXiv:2409.12191},
  eprint = {2409.12191},
  primaryclass = {cs.CV},
  publisher = {arXiv},
  doi = {10.48550/arXiv.2409.12191},
  urldate = {2026-05-24},
  archiveprefix = {arXiv},
  langid = {english}
}

@misc{yangQwen2TechnicalReport2024,
  title = {Qwen2 {{Technical Report}}},
  author = {Yang, An and Yang, Baosong and Hui, Binyuan and Zheng, Bo and Yu, Bowen and Zhou, Chang and Li, Chengpeng and Li, Chengyuan and Liu, Dayiheng and Huang, Fei and Dong, Guanting and Wei, Haoran and Lin, Huan and Tang, Jialong and Wang, Jialin and Yang, Jian and Tu, Jianhong and Zhang, Jianwei and Ma, Jianxin and Yang, Jianxin and Xu, Jin and Zhou, Jingren and Bai, Jinze and He, Jinzheng and Lin, Junyang and Dang, Kai and Lu, Keming and Chen, Keqin and Yang, Kexin and Li, Mei and Xue, Mingfeng and Ni, Na and Zhang, Pei and Wang, Peng and Peng, Ru and Men, Rui and Gao, Ruize and Lin, Runji and Wang, Shijie and Bai, Shuai and Tan, Sinan and Zhu, Tianhang and Li, Tianhao and Liu, Tianyu and Ge, Wenbin and Deng, Xiaodong and Zhou, Xiaohuan and Ren, Xingzhang and Zhang, Xinyu and Wei, Xipin and Ren, Xuancheng and Liu, Xuejing and Fan, Yang and Yao, Yang and Zhang, Yichang and Wan, Yu and Chu, Yunfei and Liu, Yuqiong and Cui, Zeyu and Zhang, Zhenru and Guo, Zhifang and Fan, Zhihao},
  year = {2024},
  month = jul,
  publisher = {arXiv},
  doi = {10.48550/arXiv.2407.10671},
  urldate = {2026-05-24},
  langid = {english}
}

@article{liReadingComprehensionL12019,
  title = {Reading Comprehension in {{L1}} and {{L2}}: {{An}} Integrative Approach},
  shorttitle = {Reading Comprehension in {{L1}} and {{L2}}},
  author = {Li, Ping and Clariana, Roy B.},
  year = {2019},
  month = may,
  journal = {Journal of Neurolinguistics},
  series = {Cross-Linguistic Perspectives on Second Language Reading},
  volume = {50},
  pages = {94--105},
  issn = {0911-6044},
  doi = {10.1016/j.jneuroling.2018.03.005},
  urldate = {2026-05-24},
  langid = {english},
  lccn = {3}
}

@article{gelman2006difference,
  title={The difference between “significant” and “not significant” is not itself statistically significant},
  author={Gelman, Andrew and Stern, Hal},
  journal={The American Statistician},
  volume={60},
  number={4},
  pages={328--331},
  year={2006},
  publisher={Taylor \& Francis}
}

@article{nieuwenhuis2011erroneous,
  title={Erroneous analyses of interactions in neuroscience: a problem of significance},
  author={Nieuwenhuis, Sander and Forstmann, Birte U and Wagenmakers, Eric-Jan},
  journal={Nature neuroscience},
  volume={14},
  number={9},
  pages={1105--1107},
  year={2011},
  publisher={Nature Publishing Group US New York}
}

@article{lu2026achieving,
  title={Achieving more human brain-like vision via human {EEG} representational alignment},
  author={Lu, Zitong and Wang, Yile and Golomb, Julie D},
  journal={Communications Biology},
  volume={9},
  number={1},
  pages={463},
  year={2026},
  publisher={Nature Publishing Group UK, London}
}

@article{shao2024probing,
  title={Probing Human Visual Robustness with Neurally-Guided Deep Neural Networks},
  author={Shao, Zhenan and Ma, Linjian and Zhou, Yiqing and Zhang, Yibo Jacky and Koyejo, Sanmi and Li, Bo and Beck, Diane M},
  journal={arXiv preprint arXiv:2405.02564},
  year={2024}
}

@article{lu2025teaching,
  title={Teaching {CORnet} human {fMRI} representations for enhanced model-brain alignment},
  author={Lu, Zitong and Wang, Yile},
  journal={Cognitive Neurodynamics},
  volume={19},
  number={1},
  pages={61},
  year={2025},
  publisher={Springer}
}

\appendix
\section{Comparison of Representative Related Works}
\label{app:related-work-matrix}

Table~\ref{tab:estimand-matrix} situates the present study among representative
model--human alignment studies. It distinguishes prior work by input setting and
the quantity being estimated, because apparent LLM--VLM differences depend on
both of these design choices.

\begin{table*}[t]
\centering
\small
\setlength{\tabcolsep}{2pt}
\begin{tabularx}{\textwidth}{@{}>{\raggedright\arraybackslash}p{0.23\textwidth} >{\raggedright\arraybackslash}p{0.22\textwidth} >{\raggedright\arraybackslash}p{0.27\textwidth} Y@{}}
\toprule
\textbf{Study} & \textbf{Input setting} & \textbf{Primary estimand} & \textbf{Main conclusion} \\
\midrule
\citet{tonevaInterpretingImprovingNaturallanguage2019} & Natural text & Sequential embedding--brain mapping & Embedding--brain mapping \\
\citet{gaoIncreasingAlignmentLarge2025} & Natural reading & Fixation-linked attention encoding & Scaling / instruction effects \\
\citet{bavarescoModellingMultimodalIntegration2025} & Concept words with picture or sentence context & Concept-word VLM--LLM brain alignment & VLM $>$ LLM overall; null Idefics2--Mistral difference in sentence-only context \\
\citet{bavarescoExperientialSemanticInformation2025} & Concepts and experiential norms & Contrastive VLM vs.\ LM experiential and fMRI alignment & LM $\geq$ contrastive VLM \\
\citet{ootaInstructionTunedVideoAudioModels2025} & Video/audio with text & Multimodal instruction-tuned brain alignment & Task-dependent \\
\citet{ryskinaLanguageModelsAlign2025} & Text and concepts & Cross-modal concept consistency & LM--brain alignment; not VLM $>$ LLM \\
\citet{subramaniamRevealingVisionLanguageIntegration2024a} & Multimodal stories & Online multimodal integration & Multimodal advantage \\
This work & Text-only natural reading & Whole-cortex, process-level, and visual-strength-conditioned differences & Global equivalence; selective local effects \\
\bottomrule
\end{tabularx}
\caption{Comparison of representative model--human alignment studies. Conclusions depend jointly on the input setting and the quantity being estimated.}
\label{tab:estimand-matrix}
\end{table*}
\section{Model and Data Details}
\label{app:model-details}

\subsection{Model Pairing Design}

Table~\ref{tab:model-configs} summarizes the architecture, backbone adaptation regime, and pretraining details for the three primary ${\sim}7$B--8B LLM/VLM pairs as well as the smaller within-family Qwen2-1.5B/Qwen2-VL-2B scale-sensitivity pair. In the Qwen2 and DeepSeek pairs, the VLM extends its matched base LLM with a vision encoder via first-party full-backbone multimodal continued pretraining, providing direct within-lineage comparisons. Idefics2-8B-base likewise initializes its language backbone from Mistral-7B-v0.1, but integrates a SigLIP vision encoder and Perceiver resampler in a third-party pipeline that adapts pretrained backbone weights via LoRA while fully training newly initialized connector parameters \citep{laurenconWhatMattersWhen2024}. Comparing full-backbone continued pretraining with this cross-construction LoRA regime clarifies how backbone adaptation depth relates to representational drift and human alignment.

All model checkpoints were obtained from Hugging Face. The specific checkpoints used are:
{\sloppy
\begin{itemize}
    \item Qwen2-7B: \url{https://huggingface.co/Qwen/Qwen2-7B}
    \item Qwen2-VL-7B: \url{https://huggingface.co/Qwen/Qwen2-VL-7B}
    \item Qwen2-1.5B: \url{https://huggingface.co/Qwen/Qwen2-1.5B}
    \item Qwen2-VL-2B: \url{https://huggingface.co/Qwen/Qwen2-VL-2B}
    \item DeepSeek-LLM-7B: \url{https://huggingface.co/deepseek-ai/deepseek-llm-7b-base}
    \item DeepSeek-VL-7B: \url{https://huggingface.co/deepseek-ai/deepseek-vl-7b-base}
    \item Mistral-7B-v0.1: \url{https://huggingface.co/mistralai/Mistral-7B-v0.1}
    \item Idefics2-8B-base: \url{https://huggingface.co/HuggingFaceM4/idefics2-8b-base}
\end{itemize}
}
All models are released under permissive open-source licenses. Qwen2/Qwen2-VL, Mistral, and Idefics2 are released under Apache 2.0, and the DeepSeek pair under the MIT license. The Reading Brain dataset is available on OpenNeuro under a CC0 public-domain license, and the Lancaster Sensorimotor Norms are publicly available for research use.

\begin{table*}[t]
\centering
\footnotesize
\begin{tabularx}{\textwidth}{@{}>{\raggedright\arraybackslash}p{0.16\textwidth} >{\raggedright\arraybackslash}p{0.05\textwidth} >{\raggedright\arraybackslash}p{0.14\textwidth} >{\raggedright\arraybackslash}p{0.06\textwidth} >{\raggedright\arraybackslash}p{0.11\textwidth} >{\raggedright\arraybackslash}p{0.16\textwidth} Y@{}}
\toprule
\textbf{Model} & \textbf{Type} & \textbf{Parameters} & \textbf{Layers} & \textbf{Heads} & \textbf{Backbone regime} & \textbf{Pretraining data} \\
\midrule
Qwen2-7B & LLM & 7.6B & 28 & 28 Q; 4 KV & Base LLM & $\sim$7T text tokens \\
Qwen2-VL-7B & VLM & 7.6B LLM + 675M ViT & 28 & 28 Q; 4 KV & Full continued PT & Multimodal PT on $\sim$1.4T text/image tokens \\
DeepSeek-LLM-7B-base & LLM & 7B & 30 & 32 Q; 32 KV & Base LLM & $\sim$2T text tokens (Chinese \& English) \\
DeepSeek-VL-7B-base & VLM & 7B LLM + hybrid vision & 30 & 32 Q; 32 KV & Full continued PT & $\sim$400B VL tokens with high text retention \\
Mistral-7B-v0.1 & LLM & 7B & 32 & 32 Q; 8 KV & Base LLM & Not publicly disclosed \\
Idefics2-8B-base & VLM & 8B (7B LLM + SigLIP + connector) & 32 & 32 Q; 8 KV & LoRA on backbones + full connector & $\sim$1.5B images and $\sim$225B text tokens \\
\midrule
Qwen2-1.5B & LLM & 1.5B & 28 & 12 Q; 2 KV & Base LLM & $\sim$7T text tokens \\
Qwen2-VL-2B & VLM & 1.5B LLM + 675M ViT & 28 & 12 Q; 2 KV & Full continued PT & Multimodal PT on $\sim$1.4T text/image tokens \\
\bottomrule
\end{tabularx}
\caption{Configurations and backbone adaptation regimes of the evaluated LLMs and VLMs (PT = pretraining). The top six checkpoints form the three primary ${\sim}7$B--8B pairs; Qwen2-1.5B / Qwen2-VL-2B forms the within-family scale-sensitivity pair.}
\label{tab:model-configs}
\end{table*}

\subsection{Data Preprocessing}

We preprocessed the raw fMRI data from the Reading Brain dataset, following \citet{gaoIncreasingAlignmentLarge2025}. Preprocessing was conducted using fMRIPrep (v25.0.0) with all default parameters. Volumes exceeding framewise displacement $> .5$\,mm or standardized DVARS $> 1.5$ were flagged as motion outliers. Final resampling to MNI space and the fsaverage5 cortical surface (10,242 vertices per hemisphere) was performed in a single interpolation step using antsApplyTransforms and mri\_vol2surf with Lanczos interpolation. The resulting surface time series were stored as GIFTI files. BOLD time series were z-scored within each run. Eye-tracking data were extracted from the events files, which contain fixation onset times and word-level area-of-interest indices. Participant 21 was excluded from eye-tracking analyses due to incomplete recording, and participants 21 and 52 were excluded from fMRI analyses due to preprocessing failures, yielding final samples of 51 and 50 participants, respectively.

\section{Encoding Model Details}
\label{app:encoding-details}

\subsection{Predictor Construction and Ridge Regression}

Both the eye-movement and fMRI encoding models share the same predictor construction. For each sentence $s$, layer $j$, and attention head $h$, the strict lower triangle of the word-level self-attention matrix $A_{s,j,h} \in \mathbb{R}^{n_s \times n_s}$ (excluding the diagonal) was retained and flattened. These vectors were concatenated across all 148 sentences, producing 7,421 lower-triangular word-pair entries. For each layer $j$, vectors from all attention heads were stacked to form the predictor matrix
\begin{equation}
X_j \in \mathbb{R}^{7421 \times H_j},
\end{equation}
where $H_j$ is the number of attention heads at layer $j$. Heads were kept as separate predictors because different heads may encode different word-to-word relations. For both modalities, we fit a ridge regression:
\begin{equation}
\label{eq:ridge}
\widehat{y} = X_j \beta_j + \epsilon,
\end{equation}
where $y$ denotes either the saccade-count vector $y_{\mathrm{sacc}}^{(i)}$ or the vertex-wise BOLD response vector $y_{\mathrm{BOLD}}^{(i,v)}$. Hidden-state controls replace $X_j$ with PCA-reduced $\phi_{lm}$ features (128 components fit on the outer training split only).

For the contextual-hidden-state control, $h_l$ and $h_m$ denote mean-pooled subword embeddings for words $l$ and $m$, respectively. We construct a feature for each directed word pair as
\begin{equation}
\phi_{lm} = [h_l; h_m; h_l-h_m; h_l\odot h_m].
\end{equation}
Thus, both the attention and hidden-state channels are evaluated on the same word-pair entries; this construction does not imply that contextual embeddings cannot be used in other alignment formulations.

\subsection{fMRI BOLD Response Construction}

For each regressive fixation transition from word $l$ to word $m$ with onset time $t_{\mathrm{onset}}$ (ms), we sampled the BOLD response at the estimated hemodynamic peak:
\begin{equation}
t_{\mathrm{scan}} = \left\lceil \frac{t_{\mathrm{onset}}/1000 + 5.0}{0.4} \right\rceil,
\end{equation}
where 5.0\,s approximates the hemodynamic peak and 0.4\,s is the repetition time (TR). For each participant $i$, sentence $s$, and cortical vertex $v$, we constructed a BOLD transition matrix $B_s^{(i,v)} \in \mathbb{R}^{n_s \times n_s}$, where entry $[l,m]$ contains the BOLD response at vertex $v$ for the fixation transition from word $l$ to word $m$. The strict lower triangle was concatenated across sentences to obtain $y_{\mathrm{BOLD}}^{(i,v)} \in \mathbb{R}^{7421}$, matching the same 7,421 word-pair entries used as predictors.

\subsection{Hyperparameter Selection and Nested Cross-validation}

All eye-movement and fMRI encoding analyses used five-fold leave-one-article-out cross-validation, with article identity defining the outer folds. In each outer fold, four articles were used for model selection and fitting, and the remaining article was reserved exclusively for evaluation.

Model layer and ridge regularization strength were treated as hyperparameters and selected jointly within the outer-training data. Specifically, we performed four-fold inner leave-one-article-out cross-validation over the four outer-training articles. For each candidate layer and ridge penalty, the encoding model was fitted on three inner-training articles and evaluated on the remaining inner-validation article. Validation scores were averaged across the four inner folds, and the layer--penalty combination with the highest mean validation performance was selected. The selected configuration was then refitted using all four outer-training articles and evaluated on the held-out outer-test article.

This procedure was repeated for all five outer folds. Predictions from the five held-out articles were concatenated to produce one complete out-of-sample prediction vector for each participant and model. Eye-movement alignment was quantified using the out-of-sample ($R^2$), whereas fMRI alignment was quantified using the Pearson correlation between predicted and observed BOLD responses at each cortical vertex.

Because layer selection was fully nested within the outer-training data, the primary statistical comparisons were conducted on the resulting participant-level nested-CV scores rather than separately at each model layer. We therefore did not apply a multiple-comparison correction across layers in the primary analysis. For each outcome, the three prespecified LLM--VLM pairwise comparisons were corrected using the Benjamini--Hochberg procedure.

\subsection{Statistical Testing}

For single-model fMRI significance, a null distribution was obtained at each cortical vertex by permuting the predicted BOLD vector 10,000 times and recomputing the prediction correlation. For matched LLM/VLM comparisons, subject-level vertex-wise $r$-maps were contrasted using paired tests across participants. Global mean-$r$ comparisons were corrected using Benjamini--Hochberg FDR correction. For cortical surface localization, we used cluster-based permutation testing with 10,000 sign-flip permutations and fsaverage5 surface adjacency \citep{marisNonparametricStatisticalTesting2007}, with a cluster-forming threshold of $p = .01$ and cluster-level $\alpha = .05$.

For eye-movement alignment, matched LLM/VLM pairs were compared using two-sided paired $t$-tests across subjects on the participant-level nested-CV scores, with Benjamini--Hochberg FDR correction across the three prespecified model-pair contrasts.

\subsection{Software and Implementation}

All analyses were implemented in Python using PyTorch 2.10, MNE 1.61, and SciPy 1.15.3. Attention extraction used Hugging Face Transformers with eager attention implementation enabled.

\section{Visualness Scoring and Modulation Details}
\label{app:visualness-details}

\subsection{Visual Strength Distribution}

\begin{figure}[ht]
    \centering
    \includegraphics[width=\columnwidth]{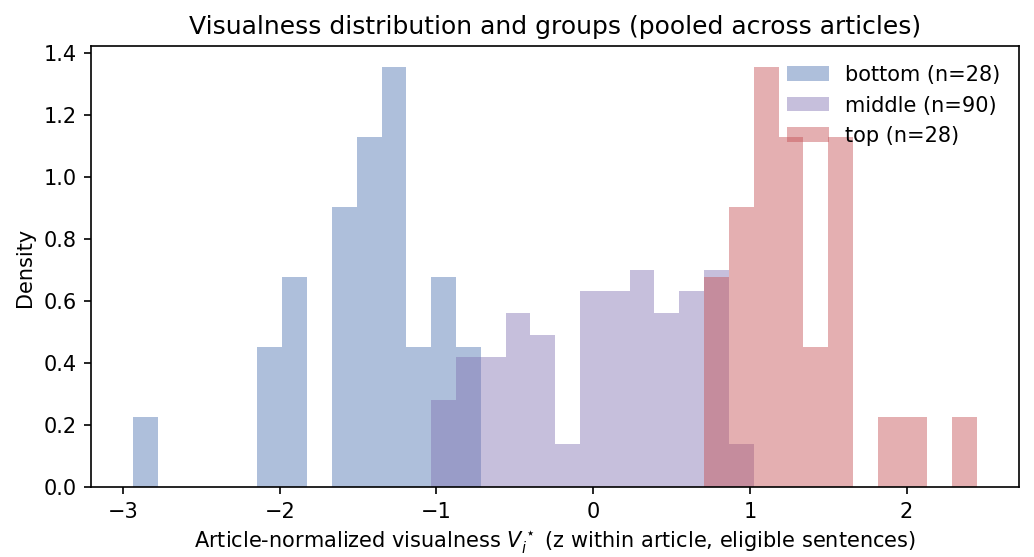}
    \caption{Distribution of sentence-level visual strength scores,
    computed from the Vision dimension of the Lancaster Sensorimotor
    Norms. Shaded regions mark the within-article top and bottom 20\%
    used for the extremes summary in
    Figure~\ref{fig:visualness}c--d and
    Table~\ref{tab:visualness-binary-appendix}.}
    \label{fig:visualness-distribution}
\end{figure}

Figure~\ref{fig:visualness-distribution} shows the distribution of
sentence-level visual strength scores across the 143 Lancaster-scored
stimulus sentences used in the moderation analysis.

\subsection{Continuous Moderation Slopes, Specification Robustness, and Explained Variance}
\label{app:visualness-continuous}

Table~\ref{tab:visualness-continuous-appendix} reports the whole-cortex
visualness-moderation slopes $\bar\beta$ for each matched pair and the
participant-matched pair contrasts used to evaluate whether pairs differ.
Slopes are estimated from Eq.~\ref{eq:moderation} via sentence-length-weighted least squares with effective word-pair count controls ($L_s$), averaged across cortical units within each
participant, and tested across participants with one-sided sign-flip
permutation tests (10{,}000 permutations) and Holm correction across
the three within-pair tests.

To verify that positive moderation in the full-backbone pairs does not depend on observation weighting or between-article baseline shifts, we conducted two regression specification checks on the same 143 Lancaster-scored sentences: (i) unweighted ordinary least squares ($w_s = 1$) and (ii) adding article fixed effects alongside $L_s$. Slopes remained consistently positive under unweighted fitting ($\bar\beta = +.0016$ for Qwen2; $\bar\beta = +.0024$ for DeepSeek) and with article fixed effects ($\bar\beta = +.0015$ for Qwen2; $\bar\beta = +.0023$ for DeepSeek). Furthermore, nested-model comparisons with and without $V_s$ (controlling for article identity and sentence length) indicate that visual strength accounts for an incremental $\Delta R^2$ of $2.0\%$ (Qwen2) and $2.8\%$ (DeepSeek) of between-sentence alignment variance at the subject level. Thus, sentence visual strength acts as a modest, systematic moderator of inter-sentence VLM--LLM neural differences rather than a dominant source of total alignment variance.

\begin{table*}[!t]
\centering
\small
\setlength{\tabcolsep}{3pt}
\begin{tabular}{@{}l@{\extracolsep{\fill}} r l r r r r@{}}
\toprule
\textbf{Contrast} & $n$ & $n_{+}/n_{-}$ & $\bar\beta$ & 95\% CI & $d_z$ & Holm $p$ \\
\midrule
\multicolumn{7}{@{}l}{\emph{Within-pair slopes}} \\[2pt]
Qwen2            & 50 & 32\,/\,18 & $+$.0017 & [$+$.0003, $+$.0031] & $+$0.35 & .015$^{*}$ \\
DeepSeek         & 50 & 34\,/\,16 & $+$.0025 & [$+$.0006, $+$.0044] & $+$0.38 & .011$^{*}$ \\
Mistral/Idefics2 & 50 & 24\,/\,26 & $-$.0006 & [$-$.0016, $+$.0005] & $-$0.16 & .857 \\
\midrule
\multicolumn{7}{@{}l}{\emph{Participant-matched pair contrasts}} \\[2pt]
Qwen $-$ Mistral      & 50 & 28\,/\,22 & $+$.0023 & [$+$.0005, $+$.0040] & $+$0.36 & .005$^{*}$ \\
DeepSeek $-$ Mistral   & 50 & 34\,/\,16 & $+$.0031 & [$+$.0009, $+$.0053] & $+$0.40 & .005$^{*}$ \\
\bottomrule
\end{tabular}
\caption{Whole-cortex visualness-moderation slopes ($n=50$).
$\bar\beta$: mean across cortical units of the regression slope of
VLM--LLM Fisher-$z$ difference on standardized Lancaster visual
strength, controlling for sentence length (Eq.~\ref{eq:moderation}). $n_{+}/n_{-}$: participants with
positive/negative slopes. One-sided sign-flip permutation tests
(10{,}000 permutations); Holm correction across three within-pair
tests. Pair contrasts are participant-matched $\Delta\bar\beta$ values
under the same procedure, with Holm correction applied separately across
the two pair-contrast tests
\citep{gelman2006difference,nieuwenhuis2011erroneous}.
$^{*}p<.05$ after Holm correction.}
\label{tab:visualness-continuous-appendix}
\end{table*}

\subsection{Grouped High vs Low Visual Strength Control}
\label{app:visualness-extremes}

To display the moderation at the tails of the visualness distribution,
we computed the whole-cortex mean Fisher-$z$ VLM--LLM advantage
separately for sentences in the within-article top and bottom 20\% of
Lancaster visual strength, and report the binary contrast
$H = A_{\mathrm{high}} - A_{\mathrm{low}}$
(Table~\ref{tab:visualness-binary-appendix}). Bayesian evidence
(two-sided JZS, Cauchy scale 0.707) favours the alternative for Qwen2
(BF$_{01}=0.16$), is ambiguous for DeepSeek (BF$_{01}=0.66$), and
favours the null for Mistral/Idefics2 (BF$_{01}=6.34$). The sign of $H$
agrees with the continuous slope $\bar\beta$ for all three pairs.

\begin{table}[t]
\centering
\small
\setlength{\tabcolsep}{2pt}
\begin{tabular}{@{}l c c c c c c@{}}
\toprule
\textbf{Pair} & $A_{\!L}$ & $A_{\!H}$ & $H$ & $d_z$ & BF$_{01}$ & $p$ \\
\midrule
Qwen2           & $-$.0014 & $+$.0030 & $+$.0043 & 0.41 & 0.16 & .003 \\
DeepSeek        & $-$.0025 & $+$.0022 & $+$.0047 & 0.32 & 0.66 & .013 \\
Mistral/Idefics2 & $+$.0015 & $+$.0014 & $-$.0001 & $-$0.02 & 6.34 & .539 \\
\bottomrule
\end{tabular}
\caption{Visualness moderation using within-article top/bottom 20\% of Lancaster visual strength ($n=50$).
$A_{\mathrm{low}}$ and $A_{\mathrm{high}}$: mean Fisher-$z$ VLM--LLM
advantages for low- and high-visual-strength sentences. $p$-values: one-sided sign-flip
(10{,}000 permutations). BF$_{01}$: two-sided JZS Bayes factor (Cauchy
scale 0.707).}
\label{tab:visualness-binary-appendix}
\end{table}

\subsection{Threshold Sensitivity: 25/25 Split}

To test sensitivity to the 20/20 grouping threshold, we repeated the
extremes summary using a within-article top/bottom 25\% split
(Table~\ref{tab:visualness-robustness}). The direction and approximate
magnitude of $H$ are unchanged for both first-party pairs.

\begin{table}[t]
\centering
\small
\setlength{\tabcolsep}{3pt}
\begin{tabular}{@{}l c r r r@{}}
\toprule
\textbf{Pair} & $n$ & $H$ & $d_z$ & Sign-flip $p$ \\
\midrule
Qwen2 & 50 & +.003542 & 0.33 & .0113 \\
DeepSeek & 50 & +.004962 & 0.34 & .0089 \\
Mistral/Idefics2 & 50 & +.000536 & 0.07 & .3147 \\
\bottomrule
\end{tabular}
\caption{Threshold-sensitivity analysis using a within-article
top/bottom 25\% split.}
\label{tab:visualness-robustness}
\end{table}

\subsection{LLM-Guided Coverage-Recovery Sensitivity Analysis}
\label{app:llm-fallback}

The moderation analysis excludes sentences without sufficient Lancaster
coverage. To assess sensitivity to that exclusion, we ran a secondary,
exploratory coverage-recovery analysis that supplemented low-coverage
sentence scores with calibrated estimates from external judges; it does
not replace the Lancaster-only analysis.

Of the 148 stimulus sentences, 143 (96.6\%) had sufficient Lancaster
coverage, four (2.7\%) received coverage-recovery estimates, and one
(0.7\%) was excluded because it contained fewer than three content words.
The external judges (GPT-5.4-2026-03-05, Gemini-2.5-flash, and
Kimi-K2.5) were outside all evaluated LLM/VLM families. Each judge
scored the four low-coverage sentences three times on scene content,
visual detail, and imageability (0--5). Scores were calibrated within
each LOAO fold against high-coverage Lancaster scores from training
articles, then fused by the across-judge median. Prompts, the JSON
schema, and decoding settings are provided in the project repository.
Algorithm~\ref{alg:llm_visualness_fallback} summarizes the procedure.
Results are reported in Table~\ref{tab:visualness-coverage-recovery}.

\begin{table}[t]
\centering
\small
\begin{tabular}{@{}l c r r@{}}
\toprule
\textbf{Pair} & $n$ & $H$ & Sign-flip $p$ \\
\midrule
Qwen2 & 50 & +.003634 & .0099 \\
DeepSeek & 50 & +.004580 & .0139 \\
Mistral/Idefics2 & 50 & $-$.000758 & .7524 \\
\bottomrule
\end{tabular}
\caption{Exploratory coverage-recovery sensitivity analysis. The
positive direction for Qwen2 and DeepSeek is unchanged relative to the
Lancaster-only analysis. Format follows
Table~\ref{tab:visualness-binary-appendix}.}
\label{tab:visualness-coverage-recovery}
\end{table}

\begin{algorithm*}[t]
\caption{External-judge coverage recovery for low-coverage sentences}
\label{alg:llm_visualness_fallback}
\begin{algorithmic}[1]
\Require Sentences $\mathcal{S}$ by article; Lancaster visual norms;  judge set $\mathcal{M}$.
\Ensure A fixed sentence-level visual strength table.

\State Define fallback set $\mathcal{F}=\{i:  \ \kappa_i<0.75 \ \mathrm{or}\ |M_i|<3\}$.

\ForAll{sentences $i \in \mathcal{F}$}
    \ForAll{judges $m \in \mathcal{M}$}
        \State Query judge $m$ three times using the fixed rubric and JSON schema.
        \State Obtain scores in $[0,5]$ for scene content, visual detail, and imageability.
        \State Average the three sub-scores; take the median across three repeated calls.
    \EndFor
\EndFor

\ForAll{leave-one-article-out folds $f$}
    \State Let $\mathcal{T}_f$ be training articles and $\mathcal{H}_f$ the held-out article.
    \ForAll{$m \in \mathcal{M}$}
        \State Fit isotonic calibration $g_f^{(m)}$ using high-coverage sentences in $\mathcal{T}_f$.
    \EndFor
    \ForAll{$i \in \mathcal{F}\cap\mathcal{H}_f$}
        \State Set $\hat{V}_i^{\mathrm{LLM}}=\mathrm{median}_{m\in\mathcal{M}} g_f^{(m)}(q_i^{(m)})$.
    \EndFor
\EndFor

\end{algorithmic}
\end{algorithm*}

\subsubsection{LLM Judge Trigger Rule}

For each sentence $i$, let $C_i$ denote the set of content words and $M_i \subseteq C_i$ the subset matched to entries in the Lancaster Sensorimotor Norms. The norm-based sentence score was defined as
\begin{equation}
V_i^{\mathrm{dict}} = \frac{1}{|M_i|} \sum_{t \in M_i} \mathrm{Vision}(t).
\end{equation}
Lexical coverage was defined as
\begin{equation}
\kappa_i = \frac{|M_i|}{|C_i|}.
\end{equation}
A sentence was assigned to the coverage-recovery procedure if $\kappa_i < 0.75$ or $|M_i| < 3$. The resulting sensitivity-analysis score was
\begin{equation}
V_i =
\begin{cases}
V_i^{\mathrm{dict}}, & \kappa_i \geq 0.75\ \text{and}\ |M_i| \geq 3, \\
\widehat{V}_i^{\mathrm{LLM}}, & \text{otherwise}.
\end{cases}
\end{equation}

\section{Representational Change Details}
\label{app:repchange}

RSA used word-level mean-pooled hidden states to build sentence-level RDMs with Pearson correlation distance ($1 - r_{\mathrm{Pearson}}$), comparing paired models via Spearman correlation across upper-triangular RDM entries (summarized by the median across layers, with sentence bootstrap CIs). Unrelated reference: Qwen2 vs.\ Mistral RSA $=.574$. Attention JSD followed Gao's midpoint implementation of cached word-level attention; per-layer medians were Qwen2 $0.0189$, DeepSeek $0.0197$, and Mistral/Idefics2 $0.0064$ after sink removal. DeepSeek sink mass (mean attention to word~0 for rows $i\geq1$) fell from $0.791$ to $0.146$. Gao instruction-tuning per-layer JSD on the same 148 sentences ranged from $0.0001$ to $0.0113$. As a deterministic pipeline sanity check, comparing identical checkpoints on the same 148 sentences recovered $\mathrm{RSA}=1.0$ and $\mathrm{JSD}=0.0$ within floating-point precision (maximum numerical deviation $<10^{-7}$); distinct checkpoints naturally diverge from $\mathrm{RSA}=1.0$ as weights update during continued training, whether unimodal or multimodal. Layer-wise curves and bootstrap summaries are archived in the project repository. 

\section{Whole-Cortex Equivalence and Sensitivity}
\label{app:equivalence-sensitivity}

For primary whole-cortex contrasts, we report the mean VLM--LLM Fisher-$z$ difference, 95\% CI, Cohen's $d_z$, two-sided paired $p$, TOST against a working $\pm0.01$ Fisher-$z$ region, and a paired JZS Bayes factor BF$_{01}$ with Cauchy scale $0.707$. We use BF$_{01}>3$ as evidence favoring the null relative to the alternative; the equivalence region is not preregistered. We re-evaluated the contrasts under SESOI bounds $\pm0.005$, $\pm0.010$, and $\pm0.020$, and JZS Cauchy scales $0.5$, $0.707$, and $1.0$. Attention- and hidden-channel whole-cortex fMRI decisions remain Equivalent at all three bounds for the reported pairs. At Cauchy $=0.5$, BF$_{01}$ for attention Mistral/Idefics2 ($2.71$) and hidden DeepSeek ($2.69$) fall slightly below the conventional threshold of 3, so Bayesian evidence should be described as moderate and prior-sensitive rather than uniformly decisive. Full grids are archived with the analysis code.

\section{Hidden-State Alignment Comparison}
\label{app:hidden-control}

To test whether the attention-based conclusions depend on the choice of representation channel, we repeated the fMRI and saccade-count analyses using contextual-hidden-state features $\phi_{lm}$ under the same nested-CV, LOAO, and the same nested layer--ridge selection procedure.

As for whole-cortex fMRI, Table~\ref{tab:hidden-fmri} reports the hidden-state whole-cortex 
equivalence tests. All three pairs yield practical equivalence under the 
$\pm0.01$ Fisher-$z$ working region, converging with the attention-based 
results in Table~\ref{tab:fmri-equivalence}.

\begin{table}[ht]
\centering
\small
\setlength{\tabcolsep}{3pt}
\begin{tabular}{@{}l c c c@{}}
\toprule
\textbf{Pair} & $\Delta$ (95\% CI) & BF$_{01}$ & TOST \\
\midrule
Qwen2 & $+$.0011 [$-$.0022, $+$.0043] & 5.29 & Equiv. \\
DeepSeek & $-$.0011 [$-$.0032, $+$.0009] & 3.58 & Equiv. \\
Mistral/Idefics2 & $-$.0002 [$-$.0020, $+$.0017] & 6.42 & Equiv. \\
\bottomrule
\end{tabular}
\caption{Hidden-state whole-cortex fMRI equivalence. Format follows Table~\ref{tab:fmri-equivalence}.}
\label{tab:hidden-fmri}
\end{table}

For regressive saccade alignment, Table~\ref{tab:hidden-saccade} reports the hidden-state saccade-count 
results. Both first-party continued-pretraining pairs show a significant 
VLM advantage, reproducing the attention-based pattern; the 
cross-construction Mistral/Idefics2 pair does not.

\begin{table}[ht]
\centering
\small
\setlength{\tabcolsep}{3pt}
\begin{tabular}{@{}l c c c@{}}
\toprule
\textbf{Pair} & $\Delta$ (95\% CI) & $p$ & Sig. \\
\midrule
Qwen2 & $+$.0016 [$+$.0002, $+$.0029] & .021 & * \\
DeepSeek & $+$.0033 [$+$.0019, $+$.0047] & $<$.001 & *** \\
Mistral/Idefics2 & $+$.0010 [$-$.0009, $+$.0029] & .289 & n.s. \\
\bottomrule
\end{tabular}
\caption{Hidden-state saccade-count alignment. Both first-party pairs replicate Figure~\ref{fig:saccade}; Mistral/Idefics2 does not.}
\label{tab:hidden-saccade}
\end{table}

The hidden-state control converges with the attention-based primary analyses on both estimands: no whole-cortex fMRI advantage in any pair, and a selective saccade advantage in the two first-party pairs. The conclusions reported in the main text are therefore not specific to the attention feature channel.

\section{Within-Family Scale Sensitivity (Qwen2-1.5B vs.\ Qwen2-VL-2B)}
\label{app:scale-sensitivity}

To test whether the behavioral-versus-global-neural dissociation observed in Qwen2-7B / Qwen2-VL-7B depends on the ${\sim}7$B parameter scale, we evaluated a smaller matched base-model pair from the same Qwen2 lineage---Qwen2-1.5B vs.\ Qwen2-VL-2B (Table~\ref{tab:model-configs})---using identical text-only inputs, word-pair attention predictors, nested LOAO cross-validation, and statistical procedures. Table~\ref{tab:scale-sensitivity} summarizes the three primary estimands.

\begin{table}[ht]
\centering
\small
\setlength{\tabcolsep}{2.5pt}
\begin{tabular}{@{}l c c@{}}
\toprule
\textbf{Estimand} & \textbf{Estimate} & \textbf{Result} \\
\midrule
Saccades ($\Delta R^2$) & $+$.0015 ($d_z{=}.395$) & $p{=}.0207^{*}$ \\
fMRI ($\Delta z$) & $-$.0005 $[{-}.0028,{+}.0018]$ & Equiv. \\
Visual slope ($\bar\beta$) & $+$.0005 & $p{=}.266$ \\
\bottomrule
\end{tabular}
\caption{Qwen2-1.5B vs.\ Qwen2-VL-2B scale check ($\Delta$: VLM$-$LLM; $^{*}$Holm $p<.05$; Equiv.: TOST $p<.001$).}
\label{tab:scale-sensitivity}
\end{table}

Qwen2-VL-2B again showed higher regressive-saccade alignment and practical whole-cortex fMRI equivalence, whereas the visual-strength slope remained positive but nonsignificant. Thus, the core behavioral-versus-global-neural dissociation holds at a smaller Qwen2 scale, while content-dependent neural moderation appears weaker at reduced model capacity.

\end{document}